\pdfoutput=1
\documentclass[11pt]{article}

\usepackage[]{ACL}

\usepackage{times}
\usepackage{latexsym}

\usepackage[T1]{fontenc}
\usepackage[utf8]{inputenc}

\usepackage{microtype}
\usepackage{inconsolata}
\usepackage{subfigure}
\usepackage{graphicx} 
\usepackage{multirow}
\usepackage{float}
\usepackage{amsmath}
\usepackage{microtype}
\usepackage{booktabs}
\usepackage{pifont}
\usepackage{xspace}
\usepackage{stfloats}
\usepackage{hyperref}
\usepackage{listings}
\usepackage{enumitem}
\usepackage{xcolor}

\newcommand{\SPE}{SPE\xspace}

\title{Serial Position Effects of Large Language Models}

\author{Xiaobo Guo \and Soroush Vosoughi\\
        Department of Computer Science \\ Dartmouth College \\
    Hanover, New Hampshire\\
    \{xiaobo.guo.gr, soroush.vosoughi\}@dartmouth.edu
    }
  
\begin{document}
\maketitle
\begin{abstract}
    Large Language Models (LLMs) have shown remarkable capabilities in zero-shot learning applications, generating responses to queries using only pre-training information without the need for additional fine-tuning. This represents a significant departure from traditional machine learning approaches. Previous research has indicated that LLMs may exhibit serial position effects, such as primacy and recency biases, which are well-documented cognitive biases in human psychology. Our extensive testing across various tasks and models confirms the widespread occurrence of these effects, although their intensity varies. We also discovered that while carefully designed prompts can somewhat mitigate these biases, their effectiveness is inconsistent. These findings underscore the significance of serial position effects during the inference process, particularly in scenarios where there are no ground truth labels, highlighting the need for greater focus on addressing these effects in LLM applications.

\end{abstract}
\section{Introduction}

    Serial position effects (\SPE{}), including the \emph{primacy} and \emph{recency} effects, are well-documented cognitive biases in human behavior. The primacy effect suggests that individuals are more likely to recall information presented at the beginning of a sequence \cite{asch1946forming}, while the recency effect implies a similar bias towards information at the end of a sequence \cite{baddeley1993recency}. These biases, attributed to factors such as diminished attention \cite{crano1977primacy}, rehearsal strategies \cite{tan2000recency}, and memory system dynamics \cite{li2010primacy}, have been extensively studied in cognitive science.

    \begin{figure}[!hbt]
        \centering
        \includegraphics[width=0.7\columnwidth]{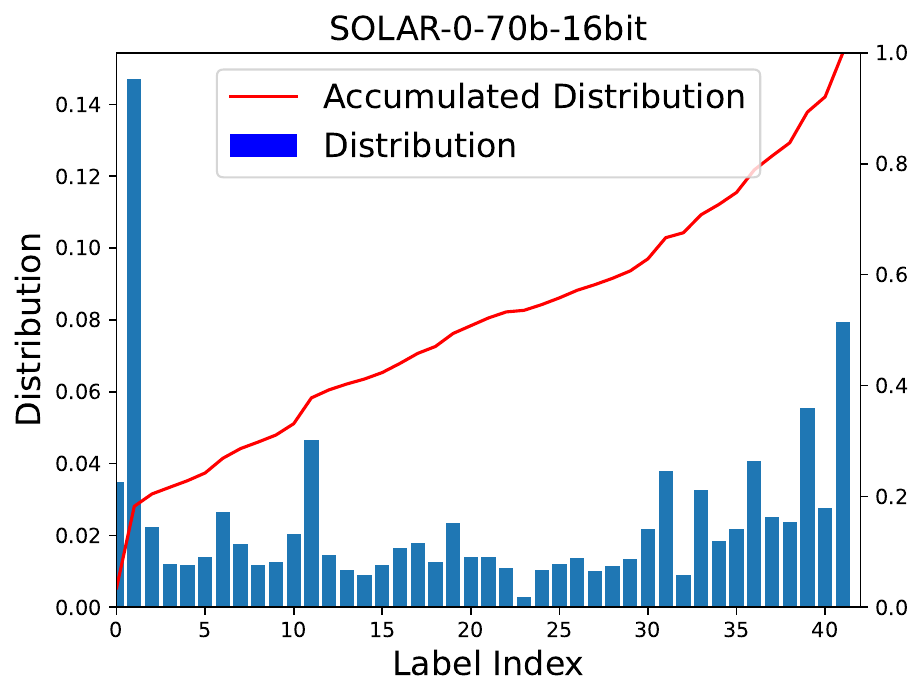}
        \caption{\SPE{} of SOLAR-0-70b-16bit: This model tends to select labels positioned at the beginning and end of a sequence more frequently. The plot illustrates the distribution of label selections across 42 labels, with the x-axis representing label positions and the y-axis the probability of selection. The red line shows the cumulative probability distribution.}
        \label{fig: serial position effect sample}
    \end{figure}

    In the context of language models like BERT, serial position effects are typically mitigated through fine-tuning, allowing the model to focus on relevant information within a context. However, with the advent of LLMs such as GPT-3.5-turbo and Llama2, known for their proficiency in zero-shot learning scenarios, there arises a need to reevaluate \SPE{} in environments where fine-tuning is not practical. These LLMs, which often do not undergo fine-tuning, may exhibit increased susceptibility to \SPE{}, thus complicating their application in real-world scenarios. This was highlighted by~\citet{zheng2023large}, who noted that LLMs could be influenced by the order of options in multiple-choice settings, a challenge exacerbated by the probabilistic processing of option identifiers (e.g., A/B/C/D).

    Further research by \citet{wang2023primacy} and \citet{zhang2023exploring} has shown that models including ChatGPT, GPT-3.5, and GPT-4 are prone to the primacy effect, a finding extended to other models like Claude-instant-1.2 by \citet{eicher2024compensatory}, indicating that these biases are not restricted to a single model family. \citet{tjuatja2023llms} suggested that techniques such as Reinforcement Learning from Human Feedback could also modulate these effects. Figure~\ref{fig: serial position effect sample} illustrates the serial position effect observed in our experiments, with the SOLAR-0-70b model as an example.

    Despite these insights, existing studies exhibit several limitations: First, they primarily examine LLMs within the GPT and Llama2 families, neglecting earlier generative models with encoder-decoder architectures such as T5 and Flan T5. Investigating these models could determine if \SPE{} are inherent to all generative models. Second, prior analyses have predominantly employed choice re-ranking methodologies. While these methods effectively demonstrate the impact of SPE, they restrict the analysis to single-label selections and fail to provide a comprehensive overview of model focus across complete inputs. Third, there is a lack of research into whether \SPE{}can be effectively mitigated during inference through straightforward interventions, such as prompt engineering and Chain-of-Thought (CoT). We focus on the inference process because LLMs are increasingly used in scenarios without a 'correct' answer, such as selecting a restaurant from a list, where fine-tuning to improve LLM performance could introduce bias. Although, in scenarios with ground truth labels, the ideal method to mitigate \SPE{} would involve enhancing the LLMs' performance, this approach falls outside the scope of this paper

    Addressing these gaps, our research expands the scope of \SPE{} investigation to include traditional LLMs and earlier encoder-decoder models like T5 and Flan T5. By conducting experiments with these encoder-decoder models, we explore whether SPEs are exclusive to decoder-only architectures or are also prevalent within broader generative models. Furthermore, our study moves beyond multiple-choice tasks to include summarization tasks, allowing for an analysis of model focus via the BERTScore correlation between source articles and generated summaries. We utilize tailored prompts designed to steer model focus and assess the impact of \SPE{} under these conditions. We also examine whether the CoT approach can guide models to thoroughly analyze all options before making decisions in multiple-choice settings. Our key findings include:
    \begin{itemize}[leftmargin=*]
    \itemsep0em 
    \item Serial position effects were consistently observed across various LLMs. While we found no significant difference in the presence of \SPE{} between encoder-decoder and decoder-only models, this suggests that \SPE{} may be a general characteristic of all generative models. However, the type and intensity of these effects vary depending on the task, which underscores the complex interplay between task characteristics and inherent biases.
    \item The use of carefully crafted prompts, including CoT, has demonstrated potential in moderating primacy and recency effects, though the success rate varies. This variability highlights the significant impact of multiple factors, such as task specifics, model selection, and prompt design.
    \item Experiments with and without prompts reveal the pervasive influence of \SPE{} and its challenging nature. These findings emphasize the need for more focused research on \SPE{}, particularly in scenarios where there are no ground truth labels to improve LLMs' '' accuracy ''.
    \end{itemize}
    
    Overall, our study contributes to the discourse on serial position effects in LLMs, elucidating their widespread occurrence, the complexities of controlling them through prompt design, and their unpredictability. These insights are crucial for effectively navigating the practical deployment of LLMs, particularly in scenarios involving complex inputs and multiple-choice questions, and provide a valuable reference for future research and application.

\section{Related Work}
    A growing body of research explores how LLMs respond to variations in prompt construction, including but not limited to permutations in multiple-choice questions~\cite{zheng2023large,zheng2023large2,pezeshkpour2023large,zhang2023exploring}, the order of in-context examples~\cite{lu2021fantastically,sclar2023quantifying}, and adversarial prompts~\cite{maus2023black,zou2023universal}. While these studies recognize the sensitivity of LLMs to prompt permutations, they primarily treat this phenomenon as a challenge in prompt engineering, rather than as a manifestation of human-like behavioral and cognitive biases.

    The comparison between LLM behavior and human cognition suggests that LLMs' sensitivity to the order of prompts may extend beyond mere engineering challenges and relate to the fundamental nature of their attention mechanisms. Research indicates various impacts of serial position effects on LLMs: \citet{wang2023primacy} finds that ChatGPT is influenced by the primacy effect; \citet{janik2023aspects} notes the recency effect in GPT-J; and \citet{zhang2023exploring} observes both effects in ChatGPT, GPT-3.5, and GPT-4, with a dominant primacy effect. Furthermore, \citet{eicher2024compensatory} extends these findings to Claude-instant-1.2, illustrating that \SPE{} is not confined to any single model family. \citet{tjuatja2023llms} posits that RLHF significantly contributes to \SPE{}.

   Previous research primarily focused on decoder-only LLMs and multiple-choice tasks. In our study, we broaden the experimental scope to encompass encoder-decoder models like T5 and FlanT5, investigating whether serial position effects (\SPE{}) impact these generative models from their inception. We also extend our inquiry to summarization tasks, hypothesizing that the more pronounced primacy effect may mask the presence of recency effects in many scenarios. Furthermore, we analyze the impact of various prompt designs as well as Cot on \SPE{}, enhancing our insight into how LLMs process and prioritize information.

\section{Models and Datasets}
    In our research, we strategically selected a diverse array of LLMs spanning various model families and sizes, including the GPT-family, Llama2-family, and T5-family. This selection encompasses both closed-source and open-source LLMs, as well as an early-stage generative model renowned for its versatility across multiple tasks. Below, we detail the composition of each model family and the rationale for their inclusion: 
    
    \noindent \textbf{GPT-family}: This family includes GPT-3.5-Turbo-0613 (GPT3.5-0613), GPT-3.5-Turbo-1106 (GPT3.5-1106), GPT-3.5-Turbo-0125 (GPT3.5-0125), and GPT-4-preview-0125 (GPT4-0125). The diversity within the GPT-3.5-Turbo models allows us to explore how \SPE{} might evolve across different versions, while GPT-4 offers a comparative analysis against the change of GPT models. 
    
    \noindent \textbf{Llama2-family}: This family consists of Llama2-7b-chat (Llama2-7b), Llama2-13b-chat (Llama2-13b), and Llama2-70b-chat (Llama2-70b). These models are fine-tuned specifically for dialogue applications, aligning with our experiment's dialogic format. Additionally, we include SOLAR-0-70b-16bit\footnote{https://huggingface.co/upstage/SOLAR-0-70b-16bit} (Solar-70b), which is a variant of Llama2-70b instruction fine-tuned by SOLAR and SOLAR-10.7B-Instruct-v1.0~\cite{kim2023solar,kim2024sdpo} (Solar-11b), which adapts Mistral 7B's weights to the Llama2 architecture with further fine-tuning for dialogues. This selection enables an examination of model size and fine-tuning techniques on \SPE{}. 

    \noindent \textbf{T5-family}: This family includes the T5~\cite{2020t5} and Flan-T5~\cite{Flan} models, both in 3B (T5-3b and FlanT5-3b) and 11B (T5-11b and FlanT5-11b) versions. T5 operates as an encoder-decoder model, pre-trained on a blend of unsupervised and supervised tasks, all formatted as text-to-text conversions. Flan-T5, an extension of T5, undergoes instruction fine-tuning across over 1,000 tasks, significantly enhancing its performance. The inclusion of both models in varying sizes allows us to assess the impact of model size and instructional tuning on \SPE{}.

    \subsection{Datasets}
    Our experimental framework incorporates both classification tasks in a multiple-choice format and summarization tasks using curated datasets. These datasets are designed to test various aspects of language understanding and generation.
    
    \paragraph{Classification Datasets}
    We align our classification experiments with established benchmarks~\cite{wang2023primacy}, focusing on relation extraction, intent detection, and emotion identification. We employ the following datasets:

    \noindent  \textbf{Banking77}~\cite{casanueva2020efficient}: This dataset is utilized for intent detection within the banking sector and contains 10,003 customer service queries, each annotated with one of 77 intents.
    
      \noindent  \textbf{GoEmotions}~\cite{demszky2020goemotions}: Consists of 58,000 Reddit comments, annotated for 27 distinct emotional states plus a neutral category, providing a comprehensive resource for emotion identification.
      
       \noindent  \textbf{MASSIVE}~\cite{fitzgerald2023massive}: A large dataset for Natural Language Understanding that includes intent prediction, with annotations covering 60 different intents.
       
      \noindent  \textbf{TACRED}~\cite{zhang2017position}: A key dataset for relation extraction that includes over 106,000 instances annotated across 42 relationship types, representing one of the largest datasets in its field.
      
     \noindent  \textbf{RE-TACRED}~\cite{stoica2021re}: An improved version of TACRED with enhanced label accuracy, where over 25\% of the entries have been re-annotated.

    To ensure robust and generalizable results, we randomly select 3,000 samples from each dataset for testing sets. This random selection process is designed to eliminate any potential bias in dataset sampling and ensure that our findings are representative of broader model capabilities.
    
    \paragraph{Summarization Datasets}
    For summarization tasks, we have compiled small sets of news articles from CNN, categorized by article count into Summ5, Summ10, and Summ20 datasets, based on whether they contain 5, 10, or 20 articles respectively. Each article is truncated to its first two paragraphs to standardize the content length to between 40 and 70 words. The articles are then randomly reordered and aggregated to create unique summarization challenges.
    
    Specifically, we generated 120 samples from the Summ5 dataset, and 1,000 samples each from the Summ10 and Summ20 datasets. This method allows us to assess the summarization capabilities of LLMs under varied and controlled conditions. Detailed listings of the news articles used in these summarization tasks are available in Appendix~\ref{Appendix: summarization datasets}.

\section{Experiment Settings}

    \subsection{Label Shuffling Experiments}
\begin{figure}[!hbt]
        \centering
        \includegraphics[width=1.0\columnwidth]{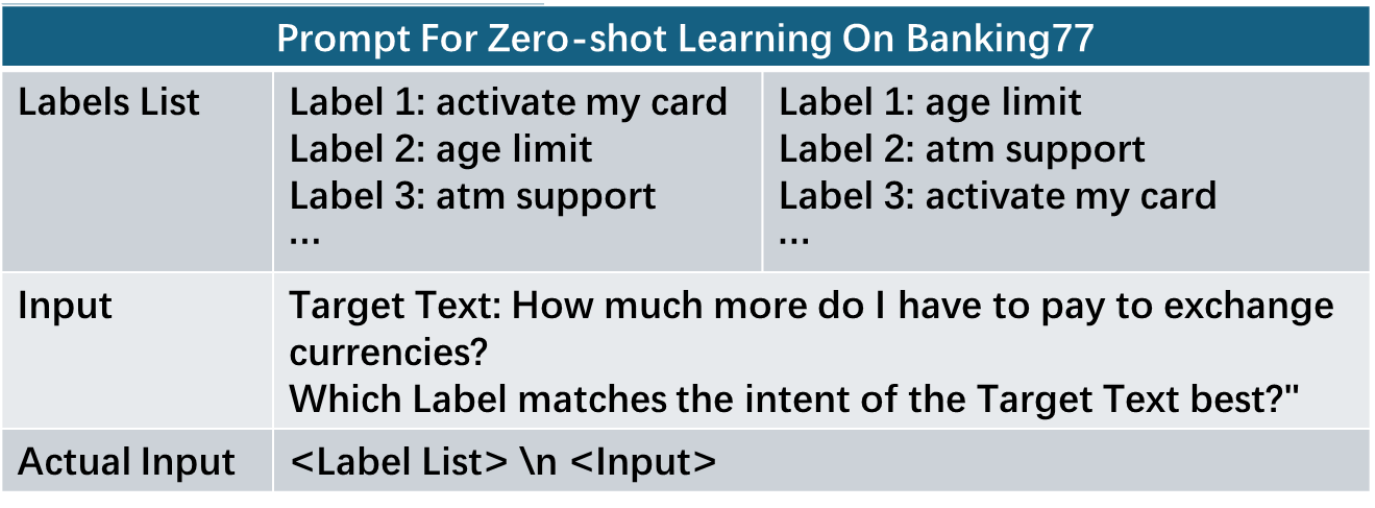}
        \caption{Examples from the Banking77 dataset where the input remains the same, but the labels are shuffled.}
        \label{fig: classification sample}
    \end{figure}

    In our classification tasks, models select the correct label from a list of shuffled candidates, a methodology adapted from prior work~\cite{wang2023primacy}. For each sample, labels within the prompt are randomized to produce two variants with identical input texts but different label orders, as illustrated in Figure~\ref{fig: classification sample}. This experimental setup allows us to examine the presence and magnitude of \SPE{}. Appendix~\ref{Appendix: prompts for label shuffling experiments} shows the prompts for these tasks.
    
    To identify \SPE{}, we analyze the cumulative distribution of predicted labels across our experiments. We define a \textbf{primacy effect} (P) as occurring when the first third of the labels account for more than 40\% of the predictions. Similarly, a \textbf{recency effect} (R) is identified if the last third exceeds 40\% of predictions. We also define a \textbf{middle effect} (M) when the middle third of labels reaches this threshold. The absence of these conditions is labeled as \textbf{no \SPE{}} (N). Multiple \SPE{} types can coexist within the same distribution. These thresholds are derived from empirical observations, given the lack of standardized criteria for \SPE{} in the current literature.

    To quantify the magnitude of these effects, we use the Jensen–Shannon divergence (JS)~\cite{menendez1997jensen}. This measure compares the predicted label distribution ($\hat{P}$) against a reference distribution ($R$). The magnitude of the serial position effects (SPEM) is then calculated as $\text{SPEM} = JS(\hat{P}||R)$. This statistical approach allows us to assess how significantly the label distribution deviates from expected norms due to \SPE{}.

\begin{figure*}[!hbt]
    \centering
    \includegraphics[width=1.99\columnwidth]{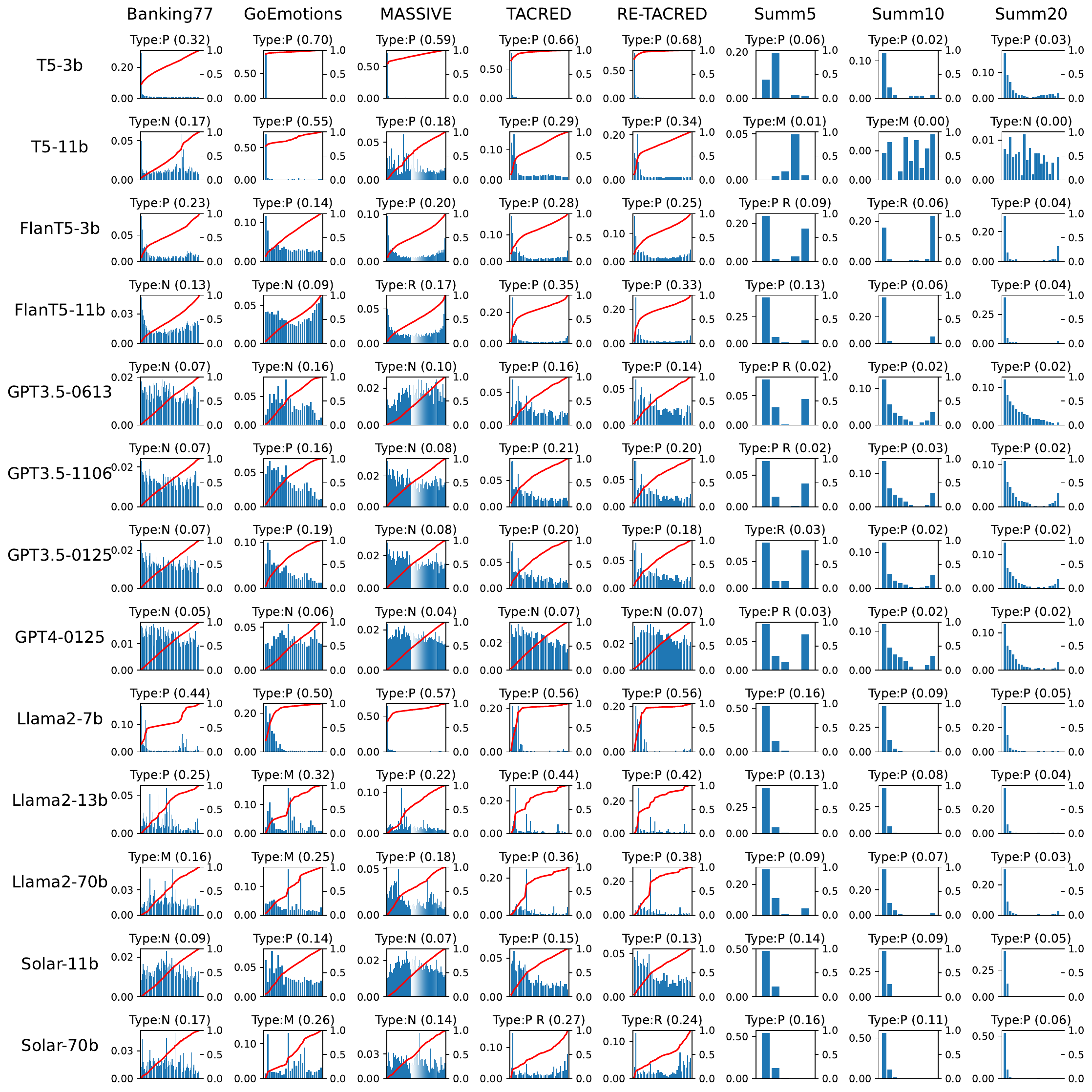}
    \caption{Distributions and cumulative distributions of predicted labels for each task across all models, with the type of \SPE{} indicated at the top of each figure and the SPEM noted in brackets. The x-axes represent the position of the labels or articles for summarization tasks. The y-axes indicate the difference in BERTScores for summarization tasks and the probability of label selection for other tasks. Red lines illustrate the cumulative probabilities.}
    \label{fig: serial position effects for plain}
\end{figure*}

    \subsection{Summarization Experiments}
    In our summarization experiments, we utilize a collection of news articles from CNN that are reordered for each experiment. Each model's task is to summarize this shuffled article set concisely. For details on the prompts used in these experiments, refer to Appendix~\ref{Appendix: prompts for summarization experiments}. To determine which articles were most focused on by the models, we calculate the normalized BERTScore for each article relative to its generated summary. This method allows us to deduce how focus is allocated across the articles, considering their positions within the input sequence. We selected BERTScore for two main reasons: 1) Our experiments are centered on the generated text as perceived by users, making BERTScore, which quantifies the similarity between source articles and generated text, particularly relevant; 2) Some LLMs, such as those in the GPT series, do not provide access to attention distributions, prompting us to employ a framework that is applicable across various models, including those from commercial entities like GPT and Gemini.
    
    Additionally, we adopt a quantitative approach to define the type of Serial Position Effect (\SPE{}) observed. We first calculate the difference in BERTScore between each article's position and the position with the lowest BERTScore. These differences are then aggregated to establish a baseline for identifying \SPE{} types, analogous to the methodology used in our label shuffling experiments. For instance, if the aggregated difference for the first third of the articles constitutes more than 40\% of the total aggregated difference, we classify this as a primacy effect. The SPEM is quantified as the mean of the absolute differences between the BERTScore at each article position and the overall average BERTScore.

\section{Influence of the Serial Position Effects}
        Figure~\ref{fig: serial position effects for plain} illustrates the distribution and cumulative distribution of predicted labels from our label shuffling experiments, as well as the BERTScore differences from summarization tasks across all models. Each chart includes annotations for the type and magnitude of \SPE{} (Type and SPEM).

        Analysis of Figure~\ref{fig: serial position effects for plain} reveals that the primacy effect is the most common across all models and tasks, appearing in 73 out of 104 instances. This effect consistently manifests across various tasks, particularly in models such as T5-3b and Llama2-7b-chat. Conversely, variations in the manifestation of \SPE{} are observed among different models; notably, GPT-4-0125-preview did not show any \SPE{} in label shuffling tasks, although it was present in summarization tasks. This inconsistency could be attributed to GPT-4's enhanced accuracy or specific design features that affects the model attention.

        Further examination shows that the recency effect is more pronounced in summarization tasks, albeit generally less dominant than the primacy effect when both are present. This trend suggests that in inherently classification-oriented tasks such as label shuffling, the recency effect tends to be overshadowed by the more prominent primacy effect.

        Specifically, in the Summ5 setup, which only involves 5 articles, the recency effect is clearly visible. However, in the Summ20 scenario with 20 articles, the effect, while still present, is too subtle to meet our identification criteria. This indicates that as the length of the prompt increases, the focus of attention shifts significantly towards the beginning of the input. Thus, it is essential to carefully manage prompt length to effectively distribute model attention, prioritizing content over introductory or guiding information at the start of the prompt. In contrast, the middle sections of prompts generally receive the least attention, highlighting the importance of understanding how information is prioritized and processed, especially in tasks that involve extensive information synthesis.

        Unique observations were made in the GoEmotions task using Llama2 family models (Llama2-13b-chat, Llama2-70b-chat, and SOLAR-0-70b-16bit), where a middle effect was detected. This scenario, wherein models predominantly focus on the central options, is not commonly reported in cognitive science research on human behavior. We hypothesize that the pre-training phase, rather than fine-tuning or RLHF processes, significantly influences the emergence of this specific \SPE{}. This aligns with previous findings \cite{janik2023aspects}, which suggest a critical role for pre-training in how LLMs prioritize and process input.

\section{Potential Methods for Mitigating Serial Position Effects}

    As demonstrated in Figure~\ref{fig: serial position effects for plain}, \SPE{} are widespread across LLMs, affecting models with various training and fine-tuning methodologies across all tasks. This section evaluates methods that could mitigate \SPE{} during inference. We specifically explore techniques such as prompting and CoT for this purpose. Although approaches like few-shot learning and fine-tuning could potentially reduce \SPE{}, they risk introducing bias in scenarios lacking ground truth due to the particular examples used. Similarly, we avoid employing self-refinement because it necessitates LLMs to generate feedback, which may further introduce bias depending on the nature of the feedback required.
    \subsection{Prompting Experiments}
        \subsubsection{Experiments with Prompts}
            These experiments are designed to assess the potential of specific prompts to alter the manifestation of \SPE{} within LLMs. In addition to a standard baseline prompt (\textbf{Plain}) which is shown in Appendix~\ref{Appendix: prompts for label shuffling experiments} and Appendix~\ref{Appendix: prompts for summarization experiments}, we have crafted six additional prompts aimed at directing the model’s foucs towards different segments of the label list: 
                
            \noindent  \textbf{Last1} and \textbf{Last2}: These prompts target the last third of the label list, encouraging models to focus on the later entries.
             
            \noindent  \textbf{Middle1} and \textbf{Middle2}: These aim at the middle third of the label/article list, to assess if central focus can convert the natural primacy or recency biases to the middle effect which is not observed in human behavior.
            
            \noindent  \textbf{Average1} and \textbf{Average2}: Designed to promote an even distribution of attention across all parts of the label/articles list.

\begin{figure}[!hbt]
                \centering
                \includegraphics[width=1.0\columnwidth]{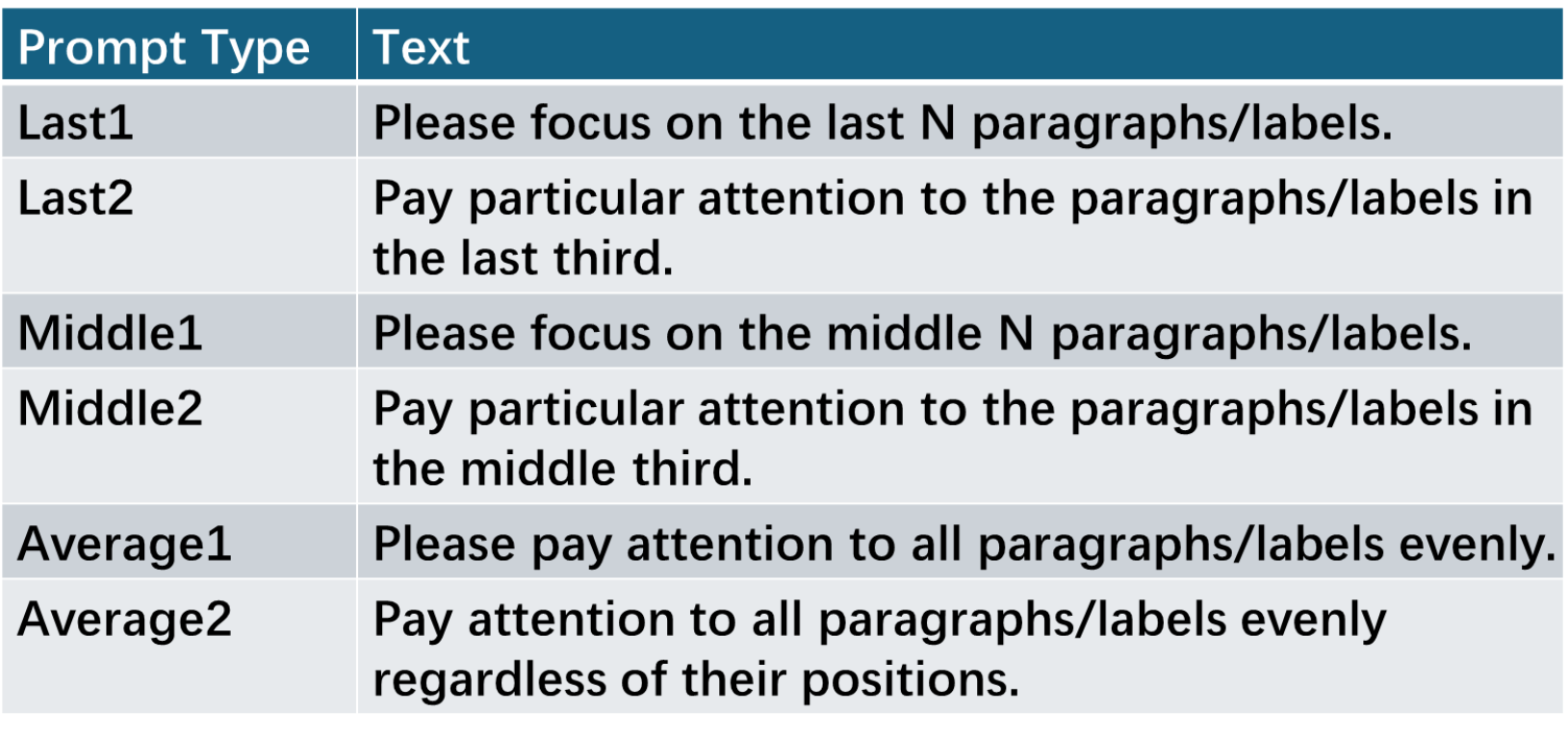}
                \caption{Illustration of the various prompts used to direct model attention to specific parts of the input. ``N'' represents the number of labels constituting one-third of the total list. }
                \label{fig: guiding prompts}
            \end{figure}

            \noindent Each of these modified prompts incorporates a directive sentence that builds upon the \textbf{Plain} prompt, explicitly guiding the model’s focus as illustrated in Figure~\ref{fig: guiding prompts}.
            
            To rigorously assess the impact of prompt design on \SPE{} in our experiments, each prompt was evaluated in comparison to the 'Plain' prompt to determine how effectively it could modify the distribution of predicted labels. We measured the following two metrics:
        
            \noindent  \textbf{Shift in \SPE{} Type}: We quantified the number of instances where the type of SPE (primacy, recency, middle, or no) shifted due to the influence of the prompt compared to the Plain prompt. This metric helped us understand the direct influence of each prompt in altering the cognitive biases demonstrated by the models. Considering that it's possible for the shift of \SPE{} type not to follow the instruction of the prompts (e.g., the model is asked to focus on the Last third while it focuses on the middle third), we further split the shift of \SPE{} type into following/not following the prompt.
            
            \noindent  \textbf{Change in SPE Magnitude for Unshifted Samples}: For samples where the \SPE{} type remained unchanged, we calculated the mean change in the SPEM. This measurement allowed us to assess the subtler effects of prompts on the intensity of existing \SPE{}, even when the type did not shift.
        
        \subsubsection{Influence of the Prompt Design}
            Table~\ref{tab: serial position effects for various prompts} presents the results of our prompting experiments, comparing the impact of various custom-designed prompts to the standard \textbf{Plain} prompt. Detailed figures for all distributions are included in the Appendix~\ref{Appendix: influence of prompts}.
    
            \begin{table*}[!hbt]
    \setlength{\tabcolsep}{0.5mm}{}
    \centering
    \small
    \begin{tabular}{l|ccc|ccc|ccc|ccc|ccc|ccc|ccc|ccc}
    \hline
\multicolumn{1}{c}{Model} & \multicolumn{3}{c}{Banking77} & \multicolumn{3}{c}{GoEmotions} & \multicolumn{3}{c}{MASSIVE} & \multicolumn{3}{c}{TACRED} & \multicolumn{3}{c}{RE-TACRED} & \multicolumn{3}{c}{Summ5} & \multicolumn{3}{c}{Summ10} & \multicolumn{3}{c}{Summ20} \\ \hline
                          & F      & NF      & $\Delta$      & F       & NF      & $\Delta$      & F      & NF     & $\Delta$     & F     & NF     & $\Delta$     & F      & NF      & $\Delta$      & F     & NF     & $\Delta$    & F     & NF     & $\Delta$     & F     & NF     & $\Delta$     \\ \hline
T5-3b                     & 0      & 0       & -0.10      & 0       & 0       & -0.02      & 0      & 0      & -0.02     & 0     & 0      & -0.04     & 0      & 0       & -0.04      & 0     & 0      & -0.04    & 0     & 0      & -0.01     & 0     & 0      & 0.00      \\
T5-11b                    & 0      & 0       & -0.01      & 0       & 0       & -0.13      & 0      & 0      & -0.01     & 0     & 0      & -0.02     & 0      & 0       & -0.02      & 2     & 1      & -0.01    & 2     & 2      & 0.00      & 1     & 2      & 0.00      \\
FlanT5-3b                 & 0      & 0       & -0.02      & 1       & 3       & -0.03      & 2      & 1      & -0.03     & 0     & 0      & -0.03     & 0      & 0       & -0.05      & 0     & 1      & -0.03    & 1     & 5      & N/A       & 0     & 0      & -0.01     \\
FlanT5-11b                & 0      & 0       & -0.02      & 0       & 0       & 0.01       & 2      & 3      & -0.05     & 0     & 0      & -0.09     & 0      & 0       & -0.10      & 0     & 0      & 0.00     & 0     & 0      & 0.00      & 0     & 0      & 0.00      \\
GPT3.5-0613               & 0      & 0       & 0.01       & 1       & 0       & -0.01      & 1      & 0      & 0.01      & 0     & 0      & 0.02      & 0      & 0       & 0.02       & 1     & 2      & 0.00     & 1     & 0      & 0.01      & 1     & 0      & 0.00      \\
GPT3.5-1106               & 0      & 0       & 0.00       & 0       & 1       & -0.01      & 0      & 1      & 0.00      & 0     & 0      & -0.01     & 0      & 0       & -0.02      & 1     & 2      & 0.00     & 1     & 0      & 0.01      & 1     & 0      & 0.01      \\
GPT3.5-0125               & 0      & 0       & 0.00       & 0       & 1       & -0.02      & 0      & 0      & 0.00      & 0     & 0      & -0.03     & 0      & 0       & -0.02      & 0     & 4      & 0.06     & 0     & 1      & 0.01      & 1     & 0      & 0.00      \\
GPT4-0125                 & 0      & 0       & 0.01       & 1       & 0       & 0.03       & 1      & 0      & 0.01      & 0     & 0      & 0.00      & 0      & 0       & 0.01       & 3     & 1      & 0.00     & 2     & 0      & 0.01      & 1     & 0      & 0.01      \\
Llama2-7b                 & 2      & 0       & 0.02       & 0       & 1       & -0.07      & 0      & 0      & -0.03     & 0     & 0      & 0.00      & 0      & 0       & -0.02      & 1     & 0      & -0.02    & 1     & 0      & -0.01     & 1     & 0      & -0.01     \\
Llama2-13b                & 0      & 2       & 0.12       & 2       & 0       & 0.01       & 0      & 1      & 0.15      & 1     & 1      & 0.03      & 1      & 1       & 0.03       & 1     & 0      & 0.00     & 1     & 0      & 0.00      & 0     & 0      & 0.00      \\
Llama2-70b                & 2      & 0       & 0.05       & 0       & 0       & 0.02       & 2      & 0      & 0.00      & 2     & 0      & 0.02      & 2      & 0       & 0.01       & 2     & 0      & 0.00     & 1     & 0      & 0.00      & 0     & 0      & 0.00      \\
Solar-11b                 & 1      & 0       & 0.01       & 0       & 1       & 0.00       & 0      & 0      & 0.01      & 0     & 1      & 0.06      & 1      & 1       & 0.05       & 1     & 0      & 0.01     & 1     & 0      & -0.01     & 0     & 0      & 0.00      \\
Solar-70b                 & 2      & 2       & 0.01       & 2       & 0       & 0.08       & 4      & 0      & 0.00      & 3     & 3      & N/A       & 2      & 1       & 0.01       & 1     & 0      & 0.00     & 1     & 0      & -0.01     & 0     & 0      & 0.00     \\ \hline
\end{tabular}
    \caption{The number of prompts (out of six) that caused a significant change in the distribution following (F) and without following (NF) the six prompts as well as the mean change of SPEM($\Delta$) for distributions without change. A significant change is defined as a shift from one serial position effect to another.}
    \label{tab: serial position effects for various prompts}
\end{table*}

            Table~\ref{tab: serial position effects for various prompts} displays the number of prompts (out of six) that induced a significant change in the distribution, defined as a shift from one serial position effect to another. Additionally, it presents the mean change of SPEM for distributions without significant change. The results indicate that prompts can significantly influence \SPE{}, both in terms of type and magnitude, although the degree of influence varies across different models and tasks. It is also noted that when the type of \SPE{} remains unchanged, the overall impact of the prompts is relatively minimal, suggesting a ``zero-or-all'' effect where a prompt either completely alters the distribution or has a limited impact. Changes are more pronounced in summarization tasks compared to label shuffling tasks.
    
            Moreover, our analysis highlights inconsistencies in the effectiveness of prompt manipulation. For instance, although the \textbf{Last1} prompt was intended to focus attention on the last third of the input, it inadvertently increased the prominence of labels in the middle section for the FlanT5-11b model on the GoEmotions task. These results indicate potential issues with current prompting methods: 1) The modifications induced by the prompts are often insufficient to fully counteract inherent \SPE{}, occasionally merely altering the nature of the effect rather than eliminating it, as seen with Solar-70b using the \textbf{Middle2} prompt on TACRED where the expected shift to a middle focus did not meet the criteria for a recognized \SPE{}. 2) Given the sensitivity of LLMs to the nuances of prompts, there is also the possibility that the prompts do not function as intended, resulting in unpredictable adjustments in label distribution. These findings emphasize the necessity for further exploration into the usage of prompts to effectively manage \SPE{} in LLMs.

\begin{figure}[!hbt]
                \centering
                \includegraphics[width=0.95\columnwidth]{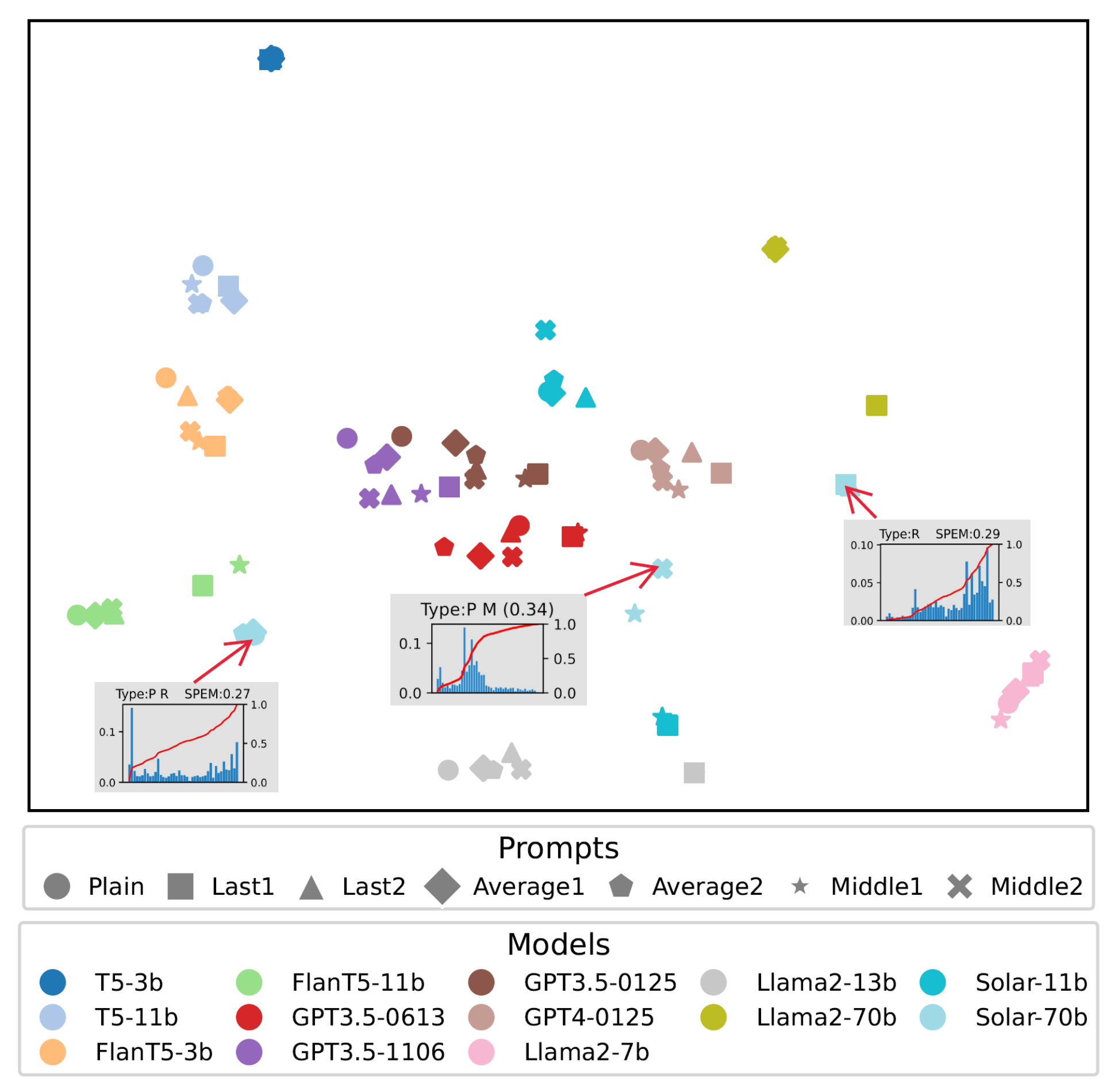}
                \caption{t-SNE visualization of label distribution for the TACRED dataset, displayed across different model and prompt combinations. Each color represents a distinct model, and various markers are used to denote different prompts.}
                \label{fig: tsne-sample}
            \end{figure}

            To delve deeper into the comparative influence of model architecture versus prompt design on the distribution of predicted labels, we conducted clustering analyses by treating each label as one dimension, thereby converting the distribution into high-dimensional data. For dimensionality reduction and visual analysis, we employed t-SNE to transform these distributions into 2-D vectors~\cite{van2008visualizing}. Figure~\ref{fig: tsne-sample} showcases these t-SNE visualizations for the TACRED dataset, which serves as a representative example among our tested datasets. Additional results are provided in Appendix~\ref{Appendix: influence of prompts}.
            
            Analyzing Figure~\ref{fig: tsne-sample}, it is observable that for certain models, such as SOLAR-70b (the distribution of different prompts are shown in Figure~\ref{fig: tsne-sample}), distributions vary with the prompts, corroborating Table~\ref{tab: serial position effects for various prompts}'s findings that prompts can modulate SPE. However, when comparing the distribution for identical prompts, distributions attributable to the same model tend to cluster together, suggesting that while prompts do affect \SPE{}, their impact is overshadowed by the intrinsic characteristics of the model itself. Beyond t-SNE visualizations, we applied clustering techniques to further quantify the relative impact of model architectures versus prompt designs on label distribution. We employed Jensen-Shannon (JS) divergence to gauge the similarity of the label distributions and subsequently used HDBSCAN for clustering the model based on the distribution similarity. Then we utilize the Adjusted Rand Index (ARI) \cite{santos2009use} to compare the alignment of the clustering groups between the cluster groups based on the model architecture and between the prompt design. This methodological approach facilitates a more nuanced understanding of how both model and prompt influence the clustering behavior of label distribution predictions.
    
            \begin{table}[!hbt]
    \centering
    \small
    \begin{tabular}{lcc}
    \hline
    Task &Model  &Prompt \\
    \hline
    Banking77&0.29  &-0.02 \\
    GoEmotions&0.20  &-0.02 \\
    MASSIVE&0.39  &-0.02 \\
    TACRED&0.35  &-0.02 \\
    RE-TACRED&0.23  &-0.02 \\
    Summ5&0.18& 0.03 \\
    Summ10&0.30  &0.06 \\
    Summ20&0.18  &-0.01 \\
    \hline
    \end{tabular}
    \caption{The Adjusted Rand Index between the clustering results and the clusters based on the model or prompt.}
    \label{tab: adj for clustering}
    \vspace{-12pt} 
\end{table}

            The data presented in Table~\ref{tab: adj for clustering} reveal that the Adjusted Rand Index between model-based clusters and prompt-based clusters are notably distinct. The ARI for prompt-based clustering groups approaches 0, suggesting that the prompt-based groupings nearly equate to random guessing in terms of their correlation with the true clustering of the distributions. In contrast, the model-based clustering achieves ARI values ranging from 0.18 to 0.39. Although these scores are far from the perfect match ARI of 1, they indicate a moderate association with the distribution of the predicted labels. This difference highlights the more substantial influence of model architecture over prompt design in shaping the label distribution outcomes in our experiments.

        \subsection{Chain-of-Thought Experiments}
        CoT is an increasingly employed technique to enhance the performance of LLMs by prompting them to generate reasoning before providing answers. In our experiments, we concentrate on classification tasks and use the following structured prompt after the \textbf{Plain} prompt: ``Generate a short explanation for your answer, analyzing all choices first. Then, choose the most suitable label from the list. Format: explanation <SEP> label.'' This prompt is specifically designed to ensure that the model considers all options thoroughly before making a decision.
        
        % \linespread{1}

\begin{table}[!hbt]
\setlength{\tabcolsep}{0.5mm}{}
\scriptsize
\begin{tabular}{lccccc}
\hline
Model       & Banking77 & GoEmotions & MASSIVE & TACRED & RE-TACRED \\ \hline
T5-3b       & -0.064    & -0.034     & -0.034  & -0.02  & -0.021    \\
T5-11b      & -0.022    & -0.157     & N/A     & -0.014 & -0.018    \\
FlanT5-3b   & -0.014    & N/A        & -0.051  & -0.073 & -0.065    \\
FlanT5-11b  & -0.003    & -0.002     & N/A     & -0.089 & -0.095    \\ \hline
GPT3.5-0613 & 0.078     & -0.018     & 0.013   & N/A    & N/A       \\
GPT3.5-1106 & 0.014     & N/A        & -0.004  & -0.029 & -0.025    \\
GPT3.5-0125 & 0.041     & -0.03      & 0       & -0.029 & -0.023    \\
GPT4-0125   & 0.019     & -0.003     & 0.002   & -0.003 & -0.005    \\ \hline
Llama2-7b   & -0.037    & -0.184     & -0.093  & -0.027 & -0.03     \\
Llama2-13b  & 0.026     & 0.004      & 0.054   & 0.03   & 0.039     \\
Llama2-70b  & N/A       & -0.073     & -0.001  & -0.027 & -0.037    \\ \hline
Solar-11b   & 0.013     & N/A        & 0.008   & 0.057  & 0.07      \\
Solar-70b   & -0.02     & N/A        & -0.031  & -0.033 & N/A       \\ \hline
\end{tabular}
\caption{The change of SPEM for distributions when utilizing CoT, ``N/A'' means there is no \SPE{} observed for this model and task. Negative means \SPE{} is mitigated compared with \textbf{Plain}.}
\label{tab: COT performance}
\end{table}
% \linespread{.9}

        Table~\ref{tab: COT performance} presents the changes in SPEM and identifies model-task pairings where \SPE{} was not observed (labeled as ``N/A''). We observe that \SPE{} is effectively mitigated across all tasks for models in the T5-Family, FlanT5-family, Solar-70b, Llama2-7b, and Llama2-70b. In the GPT family, \SPE{} is mitigated in most tasks, except for Banking77. Notably, models such as Solar-11b and Llama2-13b show divergent behavior from other models. These results indicate that CoT, with its directive to thoroughly analyze all options, can mitigate \SPE{} in LLMs for most scenarios, although it does not completely eliminate it.

\section{Conclusions}

    This study builds on previous research by investigating \SPE{} in LLMs, showing that these cognitive biases are widespread across various generative models. Our findings indicate that the prevalence of these biases is affected by model family, parameter size, and specific tasks. While strategically designed prompts can reduce these biases, their effectiveness is inconsistent across different scenarios. We recommend that future research focus on the substantial impact of \SPE{}, especially in contexts with multiple-choice formats and complex prompts where these biases are more significant.

\section{Limitations}
    \subsection{The threshold for \SPE{} Identification}
        In our experiments, the identification of \SPE{} relies on a quantitative analysis. However, given the absence of a universally recognized standard for defining \SPE{} thresholds, we have had to develop our own criteria based on empirical observations.
        
        Despite our efforts to create a standard that accurately reflects the observed distribution of attention or label probabilities, there have been instances where our predefined thresholds did not align perfectly with actual outcomes. For example, in the case of FlanT5-11b on the Banking77 task using a Plain prompt, our standard categorized this as showing no \SPE{}. Contrarily, our observations indicated that labels at the beginning and end of the list received higher probabilities than those in the middle, suggesting a potential primacy and recency effect. This discrepancy highlights the need for further refinement of our \SPE{} threshold, a task that poses significant challenges due to the complex nature of model behaviors and the subtleties of different task setups.
        
        This limitation underscores the inherent difficulty in setting a one-size-fits-all threshold for \SPE{} across various models and tasks and suggests the need for continuous evaluation and adjustment of our criteria to better capture the nuances of model responses.
        
    \subsection{Limitations Due to Language}
        In our experiments, data were exclusively sourced from English language contexts. Consequently, our conclusions regarding \SPE{} might be limited to the English context. While the probability that \SPE{} behave differently in other languages may be limited, we encourage future research to explore the impact of \SPE{} across different linguistic settings. Such studies could help determine whether the observed effects are universally applicable or if they vary significantly between languages.

    \subsection{Limitations Related to Prompt Design}
        All our experiments were conducted using prompts specifically designed for each task. Although these prompts were carefully crafted and informed by previous research, the sensitivity of LLMs to prompt nuances means that the observed \SPE{} could potentially be influenced by the specific designs used. It is possible that with better-designed prompts, the effects of \SPE{} could be more effectively manipulated.

        However, this possibility does not undermine our conclusions, as our findings also highlight the instability of using prompts to consistently control \SPE{}. This underscores the need for further exploration into prompt design as a strategy for managing cognitive biases in LLM applications.
\section{Ethic Statements}
    This study does not raise specific ethical issues as it exclusively utilizes data and models that are publicly accessible. 

\bibliography{custom}

\setcounter{table}{0}
\setcounter{figure}{0}
\renewcommand\thefigure{\Alph{section}\arabic{figure}}
\renewcommand\thetable{\Alph{section}\arabic{table}}
\clearpage
\appendix

\colorlet{punct}{red!60!black}
\definecolor{background}{HTML}{EEEEEE}
\definecolor{delim}{RGB}{20,105,176}
\colorlet{numb}{magenta!60!black}
\lstdefinelanguage{json}{
    basicstyle=\scriptsize\ttfamily,
    numbers=left,
    numberstyle=\scriptsize,
    stepnumber=1,
    numbersep=8pt,
    xleftmargin=16pt,
    showstringspaces=false,
    breaklines=true,
    frame=lines,
    backgroundcolor=\color{background},
    literate=
     *{0}{{{\color{numb}0}}}{1}
      {1}{{{\color{numb}1}}}{1}
      {2}{{{\color{numb}2}}}{1}
      {3}{{{\color{numb}3}}}{1}
      {4}{{{\color{numb}4}}}{1}
      {5}{{{\color{numb}5}}}{1}
      {6}{{{\color{numb}6}}}{1}
      {7}{{{\color{numb}7}}}{1}
      {8}{{{\color{numb}8}}}{1}
      {9}{{{\color{numb}9}}}{1}
      {:}{{{\color{punct}{:}}}}{1}
      {,}{{{\color{punct}{,}}}}{1}
      {\{}{{{\color{delim}{\{}}}}{1}
      {\}}{{{\color{delim}{\}}}}}{1}
      {[}{{{\color{delim}{[}}}}{1}
      {]}{{{\color{delim}{]}}}}{1},
}
\section{Summarization Datasets}\label{Appendix: summarization datasets}
    For the summarization dataset, we collected news articles from CNN spanning various topics. The topics include \textit{US}, \textit{World}, \textit{Politics}, \textit{Health}, \textit{Entertainment}, \textit{Style}, and \textit{Sports}. The specific news articles used are listed below:
    \begin{itemize}
        \item Warmth is set to thaw parts of the United States following frigid weekend temperatures – but the warmer air will bring a risk of ice and flooding for some states, and another crippling winter storm is set to hit portions of the Plains and South into Monday.
        \item The US carried out airstrikes in Iraq targeting facilities used by Iranian-backed militias in the country on Tuesday following repeated attacks on US forces, Defense Secretary Lloyd Austin announced in a statement. The strikes targeted three facilities used by Iranian-backed Kataib Hezbollah and other Tehran-affiliated groups in Iraq.
        \item Influencer MrBeast said on Monday that he had made more than \$250,000 from one video posted to X, in a sign of just how much major internet personalities stand to make from the social platform’s new ad revenue sharing program.
        \item Treating loneliness and social isolation may put people classified as obese at a lower risk for health complications, according to a new study. Loneliness is rampant throughout the world, but the finding is important because people with obesity experience it markedly more, the report said.
        \item As expected, the Christopher Nolan film “Oppenheimer” had a strong showing, leading Oscar contenders with 13 nominations. The fantasy film “Poor Things,” starring Emma Stone, followed with 11, while the Martin Scorsese drama “Killers of the Flower Moon” got 10 nominations.
        \item Emma Stone has just finished playing a morally complicated home-flipper in the first season of “The Curse,” and now the actor is selling her own newly-renovated Los Angeles abode.Stone has listed a two-story Westwood pied-à-terre for just under \$4 million through Sotheby’s International Reality.
        \item Normally just the unlovely places where passengers step on and off en route to somewhere more picturesque, cruise ports aren’t usually known as destinations in their own rights.But in Qatar, a regular port of call on Arabian Sea itineraries, there’s a reason to stop and stare.
        \item Embiid posted a career-high in both points and rebounds as he erupted to produce a monster stat line, ending the game with 70 points, 18 rebounds and five assists.The 29-year-old’s incredible scoring night broke the Sixers franchise record for points in a game, previously held by NBA legend Wilt Chamberlain with 68.
        \item Former South Carolina Gov. Nikki Haley gave one of her most impassioned speeches yet Tuesday as she addressed supporters in New Hampshire after CNN projected she will lose the state's Republican primary to former President Donald Trump. Haley congratulated Trump on his win but insisted she was staying in the race.
        \item Rick Slayman, the world’s first living recipient of a genetically edited pig kidney transplant, was discharged from the hospital Wednesday, two weeks after his operation, Massachusetts General Hospital said in a statement. “He is recovering well and will continue to recuperate at home with his family,” the hospital said on X, formerly Twitter.
        \item An Irish basketball team says it won’t replay the final 0.3 seconds of a playoff quarterfinal after it ended in controversial fashion, despite being instructed to do so by Basketball Ireland. The Limerick Sport Eagles beat the Portlaoise Panthers 80-78 on March 23, but have yet to play their semifinal due to confusion over the validity of their quarterfinal win.
        \item “You’ve got to get it over with, and you have to get back to normalcy. And I’m not sure that I’m loving the way they’re doing it, because you’ve got to have victory. You have to have a victory, and it’s taking a long time,” Trump said in an interview with The Hugh Hewitt Show that aired Thursday.
        \item While it was never quite clear what changes Peltz wanted Disney to implement, the 81-year-old had complained loudly about a few issues: corporate succession, “woke” entertainment, streaming strategy and profits, and a need to set up ESPN for a direct-to-streaming future.
        \item OpenAI has unveiled a new artificial intelligence tool that can mimic human voices with startling accuracy. The AI voice generator has a range of potential applications, including for accessibility services, but could also prompt concerns about misinformation and other forms of abuse.
        \item After months of legal and legislative skirmishes around the country, much of the redistricting drama of the 2024 election cycle is behind us. And it has ended pretty close to where it began: Just a handful of seats could determine which party controls the US House of Representatives, where Republicans now hold a threadbare majority.
        \item One international organized crime group makes \$50 billion a year, according to Interpol secretary-general Jurgen Stock, adding that \$2 trillion to \$3 trillion of illicit money flows through the global financial system annually. To compare, France’s economy is worth \$3.1 trillion according to the International Monetary Fund.
        \item The former head of China’s official soccer association has been sentenced to life in prison by a court in the central Chinese province of Hubei, in the latest crackdown on the country’s corruption-plagued professional football league.The ex-soccer chief, Chen Xuyuan, was jailed on Tuesday alongside multiple senior sporting executives, according to state media, following a months-long investigation.
        \item Syria and Iran blamed Israel for the airstrike that destroyed a consular building, killing Mohammed Reza Zahedi, a top commander in Iran’s elite Revolutionary Guards (IRGC), and several other officials, including another senior commander Mohammad Hadi Haji Rahimi. Israeli officials have not commented on the incident.
        \item The Dow Jones Industrial Average dropped 395 points, or 1\%, on Tuesday after declining more than 500 points at its lows. That means the blue-chip index had at one point sunk nearly 800 points during the first two days of the second quarter. On Tuesday, the S\&P500 ended the day down by 0.7\% and the Nasdaq Composite lost roughly 1\%.
        \item The burial spot (specifically, Wall B, Space C-3) is notably one row above and four spaces to the left of Monroe’s final resting place at the Pierce Brothers Westwood Village Memorial Park and Mortuary in LA. It is marginally closer to her eternal neighbor, Hefner.
    \end{itemize}
    
    The maximum normalized BERTScore between these 20 news articles is approximately 0, indicating that the embeddings of these texts are highly distinct. The diverse topics and content force the model to select specific articles as the salient information to focus on.

\section{Experiment Settings}
    \subsection{Computing Infrastructure}
    In our experiments, we utilize 4 RTX 6000 GPUs, and 64 CPU cores. The operating system of the machine is Ubuntu 20.04. Our experiments are conducted with Python 3.10. The CUDA version is 11.9 and the GPU Driver Version is 520.61.05. The details about the packages can be seen in the 'requirements.txt' of the shared codes when accepted.
    We utilize the ``Hugging Face'' implementation for the open-source language models and the official APIs of the closed-source LLMs.
    
    \subsection{Hyperparameters and Random Seed}
        In our experiments, all random seeds are set to 42 and the temperature is set to 0 for getting more deterministic results. All the other hyperparameters for the training process are set to be the default value of the package.

\section{Prompts for Experiments}
    For the label shuffling and summarization experiments, we have the following prompts which are designed based on previous work and the observation based on our experiments to ensure the model can fully understand the guides we provide. 
    \subsection{Prompts for Label Shuffling Experiments}\label{Appendix: prompts for label shuffling experiments}
    For the label shuffling experiments, we have:
        \begin{lstlisting}[language=json]
        {{#system~}}
        {{~/system}}
        {{#user~}}
            {{label_list}}
            Target Text: {{input}}
            Whiche label matches the intent of the Target Text best?
            Answer only one Label index number.
            {{#if not allow_set_assistant}}
                The label index should be: 
            {{/if}}
        {{~/user}}
        {#if allow_set_assistant}}
            {{#Assistant~}}
                The label index should be:
            {{~/Assistant}}
        {{/if}}
        \end{lstlisting}
    where \{label\_list\} is the list of label candidates, \{input\} is the text from task for classification, \{not allow\_set\_assistant\} means whether the model allows to set the text of the LLMs as part of the prompt. For LLMs no training for the chat-format such as T5 and FlanT5, we just concatenate all texts together.
    
    \subsection{Prompts for Summarization Experiments}\label{Appendix: prompts for summarization experiments}
    For the summarization experiments, we have:
        \begin{lstlisting}[language=json]
        {{#system~}}
            {{summary_instruct}ion}
        {{~/system}}
        {{#user~}}
            {{article_list}}
            {{#if not allow_set_assistant}}
                Summary: 
            {{/if}}
        {{~/user}}
        {#if allow_set_assistant}}
            {{#Assistant~}}
                Sumamry:
            {{~/Assistant}}
        {{/if}}
        \end{lstlisting}
    where \{summary\_instruction\} is the sentence for guiding the models to summary the texts, \{article\_list\} is the list of articles to be summarized split with a new line and, \{not allow\_set\_assistant\} means whether the model allows to set the text of the LLMs as part of the prompt. For LLMs no training for the chat-format such as T5 and FlanT5, we just concatenate all texts together.

    The \{summary\_instruction\} is as follow:
    \begin{itemize}
        \item  \textbf{GPT-3.5/4} Your task is to summarize the given texts. Please summarize the given texts with no more than 100 words.
        \item \textbf{Llama2-7b/13b/70b} You are an expert in summarization task. Your task is to summarize the provided paragraphs from the user.The summary should be concise. The summary should be at most 100 words.
        \item \textbf{Solar-11b/70b} Briefly summarize these paragraphs:
        \item  \textbf{T5/FlanT5} No \{summary\_instruction\}
    \end{itemize}

\section{Results}
    \subsection{The Predictability of Serial Position Effects}
        To determine the predictability of \SPE{}, our experiments focus on identifying whether it is possible to forecast the type of \SPE{} based on features such as model architecture, model size, and prompt design which are all task-irrelevant factors. In addition to model characteristics and prompt design, we examine the rate at which predicted labels changed after shuffling, which serves as an indicator of how likely a model is to alter its predictions in response to different label arrangements. We also consider model accuracy as a factor, under the hypothesis that higher accuracy should mitigate \SPE{} effects—ideally, a model with 100\% accuracy would exhibit no \SPE{} influence.
    
        These analyses are conducted exclusively with the label shuffling datasets due to the challenges of calculating accuracy and change rates in the context of summarization tasks. For each dataset, and for each type of \SPE{}, we employ logistic regression to explore potential influencing factors on the existence of \SPE{}. 

        The independent variables considered in our experiments are:
        \begin{itemize}
            \item Model Size: The number of parameters in the model.
            \item Accuracy: The accuracy of the classification task.
            \item Change Rate: The probability of the model changing its predicted label upon shuffling the labels.
            \item Model Architecture: A dummy variable representing the architecture of the model.
            \item Prompt: A dummy variable representing the specific prompt used.
        \end{itemize}
        We regard each model family as one architecture since the models in the one model family usually share the same architecture, pre-training, and fine-tuning methods. Therefore the model family includes multiple features implicitely.

        The results of these analyses for each \SPE{} type on the MASSIVE dataset are presented in Table~\ref{tab: logistic regression primacy}, Table~\ref{tab: logistic regression middle}, Table~\ref{tab: logistic regression recency}, and Table~\ref{tab: logistic regression no}. As observed, none of the independent variables significantly explain the type of \SPE{}, as indicated by the lack of significant p-values. All other tasks show similar behavior, in that the logistic regression analysis does not identify any features that significantly influenced the type of \SPE{} (with p < 0.05 in the z-test). This could be either due to our limited sample size or show that these factors are not predictive of the \SPE{} in LLMs.

        \begin{table*}[!hbt]
    \centering
    \begin{tabular}{lllllll}
    \hline
    Independent Variable & coef     & std err & z      & P\textgreater{}|z| & {[}0.025 & 0.975{]} \\ \hline
    Const                & -0.2323  & 43.147  & -0.005 & 0.996              & -84.799  & 84.335   \\
    Model Size           & -0.0345  & 0.103   & -0.334 & 0.738              & -0.237   & 0.168    \\
    Accuracy             & -7.8404  & 49.063  & -0.160 & 0.873              & -104.001 & 88.320   \\
    Change Rate          & 7.8748   & 39.741  & 0.198  & 0.843              & -70.016  & 85.765   \\
    Model GPT3.5         & 2.5102   & 20.364  & 0.123  & 0.902              & -37.402  & 42.423   \\
    Model GPT4           & -0.5330  & 66.477  & -0.008 & 0.994              & -130.826 & 129.760  \\
    Model Llama2         & 4.4770   & 9.698   & 0.462  & 0.644              & -14.530  & 23.484   \\
    Model Solar          & -10.8753 & 50.843  & -0.214 & 0.831              & -110.526 & 88.775   \\
    Model T5             & 8.9742   & 20.262  & 0.443  & 0.658              & -30.738  & 48.687   \\
    Prompt Average2      & 3.5134   & 2.740   & 1.282  & 0.200              & -1.858   & 8.884    \\
    Prompt Last1         & -6.6558  & 4.993   & -1.333 & 0.183              & -16.442  & 3.130    \\
    Prompt Last2         & -9.6005  & 5.828   & -1.647 & 0.099              & -21.023  & 1.822    \\
    Prompt Middle1       & -1.3958  & 3.747   & -0.372 & 0.710              & -8.741   & 5.949    \\
    Prompt Middle2       & 2.7370   & 2.722   & 1.005  & 0.315              & -2.598   & 8.072    \\
    Prompt Plain         & 2.5453   & 2.746   & 0.927  & 0.354              & -2.836   & 7.927    \\ \hline
    \end{tabular}
    \caption{The logistic regression analysis of the primacy effect (P) on MASSIVE.}
    \label{tab: logistic regression primacy}
\end{table*}
        \begin{table*}[!hbt]
    \centering
    \begin{tabular}{lllllll}
    \hline
    Independent Variable & coef     & std err  & z      & P\textgreater{}|z| & {[}0.025  & 0.975{]} \\ \hline
    Const                & -17.1279 & 1526.169 & -0.011 & 0.991              & -3008.364 & 2974.109 \\
    Model Size           & 0.0742   & 1.008    & 0.074  & 0.941              & -1.901    & 2.050    \\
    Accuracy             & -6.8971  & 1689.884 & -0.004 & 0.997              & -3319.009 & 3305.215 \\
    Change Rate          & -8.4425  & 769.854  & -0.011 & 0.991              & -1517.329 & 1500.444 \\
    Model GPT3.5         & -9.1627  & 793.189  & -0.012 & 0.991              & -1563.785 & 1545.460 \\
    Model GPT4           & 2.7879   & 834.178  & 0.003  & 0.997              & -1632.172 & 1637.748 \\
    Model Llama2         & -7.5794  & 1.07e+04 & -0.001 & 0.999              & -2.09e+04 & 2.09e+04 \\
    Model Solar          & 14.2664  & 761.078  & 0.019  & 0.985              & -1477.419 & 1505.952 \\
    Model T5             & -6.2258  & 3.19e+04 & -0.000 & 1.000              & -6.25e+04 & 6.24e+04 \\
    Prompt Average2      & -6.1575  & 328.056  & -0.019 & 0.985              & -649.136  & 636.821  \\
    Prompt Last1         & -6.0540  & 362.719  & -0.017 & 0.987              & -716.970  & 704.862  \\
    Prompt Last2         & -6.0941  & 355.005  & -0.017 & 0.986              & -701.891  & 689.702  \\
    Prompt Middle1       & 9.5030   & 19.830   & 0.479  & 0.632              & -29.364   & 48.370   \\
    Prompt Middle2       & 6.6370   & 28.395   & 0.234  & 0.815              & -49.016   & 62.290   \\
    Prompt Plain         & -6.1430  & 340.174  & -0.018 & 0.986              & -672.871  & 660.585  \\ \hline
    \end{tabular}
    \caption{The logistic regression analysis of the middle effect (M) on MASSIVE.}
    \label{tab: logistic regression middle}
\end{table*}
        \begin{table*}[!hbt]
    \centering
    \begin{tabular}{lllllll}
    \hline
    Independent Variable & coef    & std err & z      & P\textgreater{}|z| & {[}0.025 & 0.975{]} \\ \hline
    Const                & -2.9071 & 25.153  & -0.116 & 0.908              & -52.206  & 46.392   \\
    Model Size           & -0.0006 & 0.032   & -0.019 & 0.985              & -0.064   & 0.063    \\
    Accuracy             & 1.5068  & 28.711  & 0.052  & 0.958              & -54.766  & 57.780   \\
    Change Rate          & -2.4751 & 22.161  & -0.112 & 0.911              & -45.910  & 40.960   \\
    Model GPT3.5         & -2.8285 & 5.648   & -0.501 & 0.617              & -13.899  & 8.242    \\
    Model GPT4           & -4.7585 & 5.898   & -0.807 & 0.420              & -16.318  & 6.801    \\
    Model Llama2         & -1.3456 & 2.734   & -0.492 & 0.623              & -6.703   & 4.012    \\
    Model Solar          & -2.5062 & 1.953   & -1.283 & 0.199              & -6.333   & 1.321    \\
    Model T5             & -6.6351 & 15.061  & -0.441 & 0.660              & -36.155  & 22.885   \\
    Prompt Average2      & -4.4299 & 25.454  & -0.174 & 0.862              & -54.318  & 45.458   \\
    Prompt Last1         & 5.5788  & 3.035   & 1.838  & 0.066              & -0.369   & 11.527   \\
    Prompt Last2         & 4.2773  & 3.006   & 1.423  & 0.155              & -1.614   & 10.168   \\
    Prompt Middle1       & 2.6358  & 3.039   & 0.867  & 0.386              & -3.320   & 8.592    \\
    Prompt Middle2       & -4.3838 & 25.218  & -0.174 & 0.862              & -53.810  & 45.042   \\
    Prompt Plain         & 2.5627  & 3.048   & 0.841  & 0.400              & -3.411   & 8.537    \\ \hline
    \end{tabular}
    \caption{The logistic regression analysis of the recency effect (R) on MASSIVE.}
    \label{tab: logistic regression recency}
\end{table*}
        \begin{table*}[!hbt]
    \centering
    \begin{tabular}{lllllll}
    \hline
    Independent Variable & coef     & std err & z      & P\textgreater{}|z| & {[}0.025  & 0.975{]} \\ \hline
    Const                & 0.1887   & 46.332  & 0.004  & 0.997              & -90.620   & 90.997   \\
    Model Size           & -0.0072  & 0.033   & -0.214 & 0.830              & -0.073    & 0.058    \\
    Accuracy             & 8.0467   & 56.176  & 0.143  & 0.886              & -102.056  & 118.149  \\
    Change Rate          & -10.8739 & 34.837  & -0.312 & 0.755              & -79.152   & 57.405   \\
    Model GPT3.5         & 6.0528   & 6.680   & 0.906  & 0.365              & -7.041    & 19.146   \\
    Model GPT4           & 0.7361   & 8.396   & 0.088  & 0.930              & -15.720   & 17.192   \\
    Model Llama2         & 0.4868   & 3.551   & 0.137  & 0.891              & -6.472    & 7.446    \\
    Model Solar          & 4.1638   & 2.666   & 1.562  & 0.118              & -1.061    & 9.389    \\
    Model T5             & -4.9998  & 633.227 & -0.008 & 0.994              & -1246.102 & 1236.102 \\
    Prompt Average2      & -1.5448  & 1.693   & -0.912 & 0.362              & -4.863    & 1.774    \\
    Prompt Last1         & -2.5335  & 1.672   & -1.515 & 0.130              & -5.811    & 0.744    \\
    Prompt Last2         & 1.0264   & 1.961   & 0.523  & 0.601              & -2.818    & 4.871    \\
    Prompt Middle1       & -1.8336  & 1.687   & -1.087 & 0.277              & -5.140    & 1.473    \\
    Prompt Middle2       & -1.1593  & 1.707   & -0.679 & 0.497              & -4.505    & 2.187    \\
    Prompt Plain         & -1.4924  & 1.716   & -0.870 & 0.384              & -4.855    & 1.870    \\ \hline
    \end{tabular}
    \caption{The logistic regression analysis of the no \SPE{} (N) on MASSIVE.}
    \label{tab: logistic regression no}
\end{table*}

    \subsection{The influence of the Prompts} \label{Appendix: influence of prompts}
\begin{figure*}[!hbt]
        \centering
        \includegraphics[width=1.9\columnwidth]{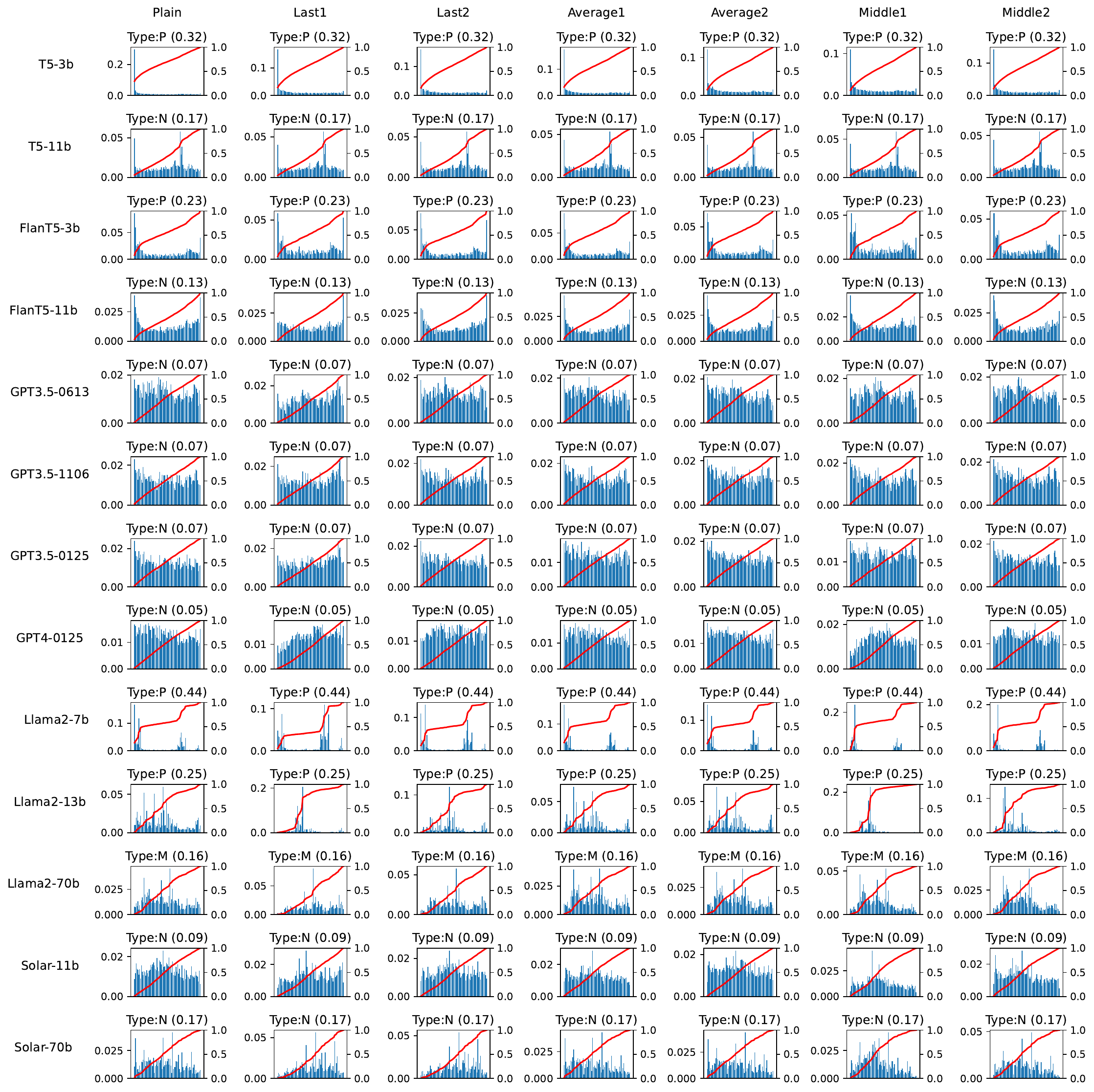}
        \caption{The label distribution of all models and prompts on the task of Banking77. The x-axis is the position of the labels. The y-axes are the probability of the label being chosen. The red line is the cumulative probability distribution. The \SPE{} type in labeled on the top of each distribution with the SPEM in the bracket.}
        \label{fig: banking77-distribution}
    \end{figure*}

\begin{figure*}[!hbt]
        \centering
        \includegraphics[width=1.9\columnwidth]{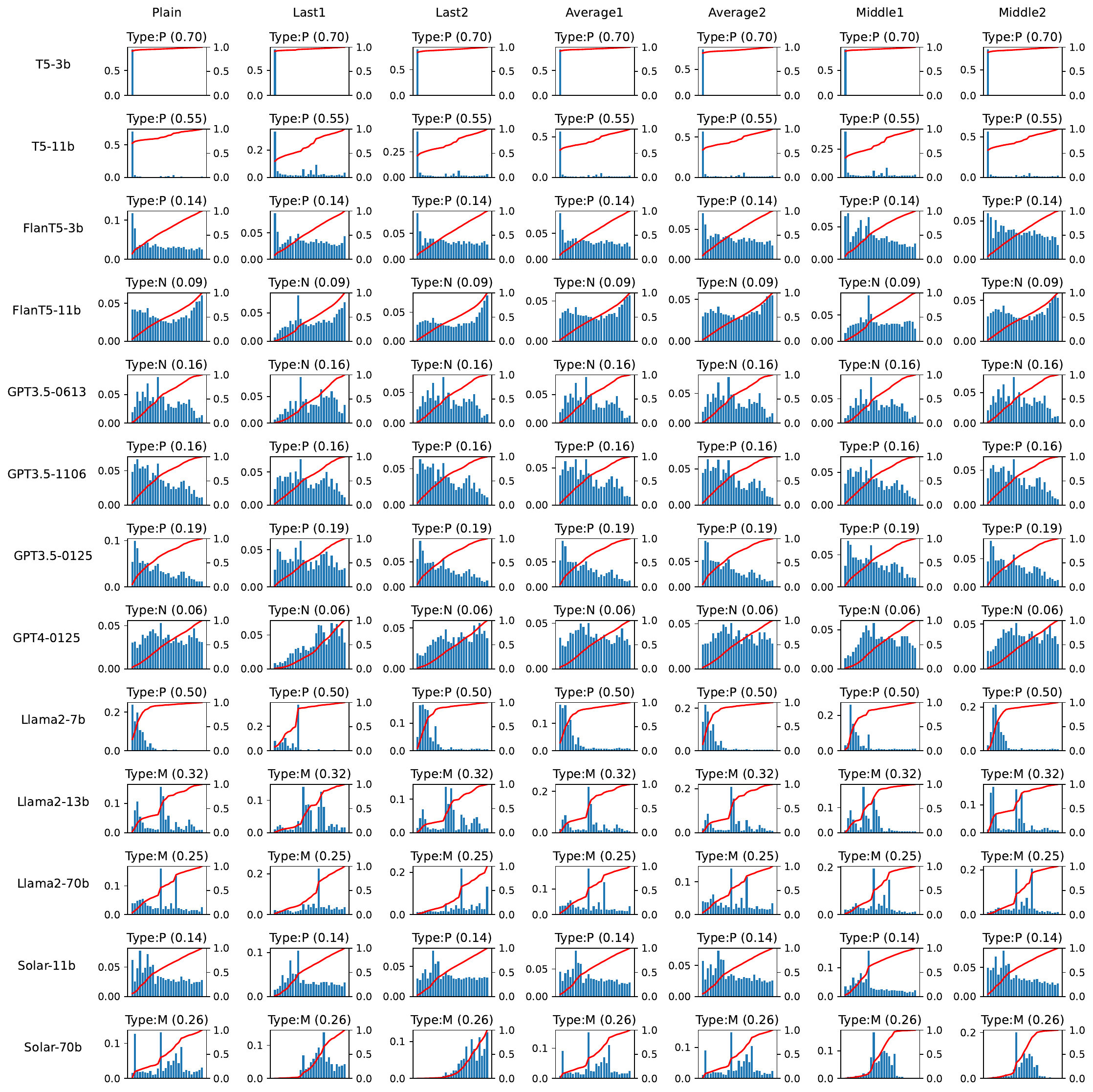}
        \caption{The label distribution of all models and prompts on the task of GoEmotions. The x-axis is the position of the labels. The y-axes are the probability of the label being chosen. The red line is the cumulative probability distribution. The \SPE{} type in labeled on the top of each distribution with the SPEM in the bracket.}
        \label{fig: goemotions-distribution}
    \end{figure*}

\begin{figure*}[!hbt]
        \centering
        \includegraphics[width=1.9\columnwidth]{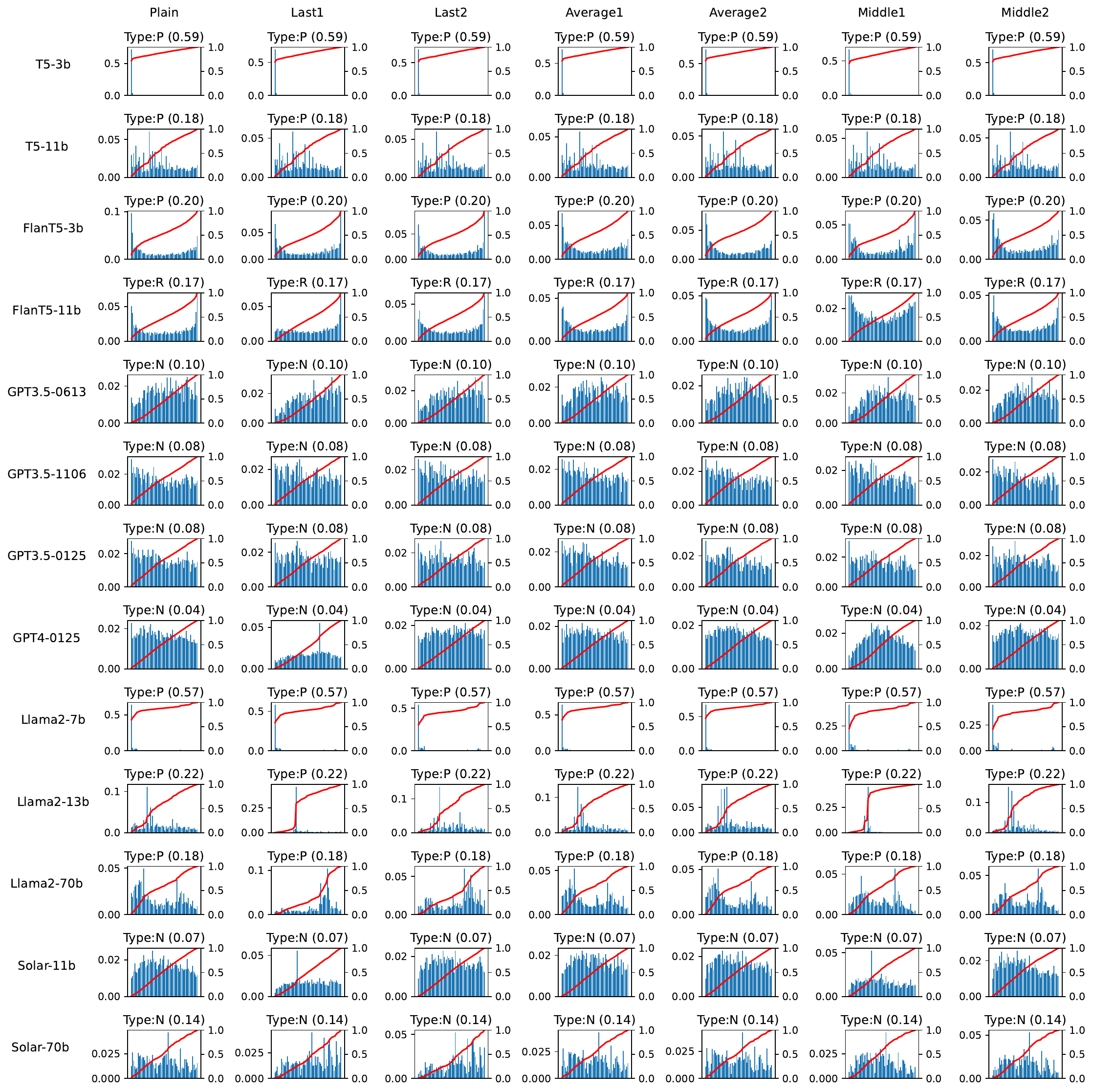}
        \caption{The label distribution of all models and prompts on the task of MASSIVE. The x-axis is the position of the labels. The y-axes are the probability of the label being chosen. The red line is the cumulative probability distribution. The \SPE{} type in labeled on the top of each distribution with the SPEM in the bracket.}
        \label{fig: massive-distribution}
    \end{figure*}

\begin{figure*}[!hbt]
        \centering
        \includegraphics[width=1.9\columnwidth]{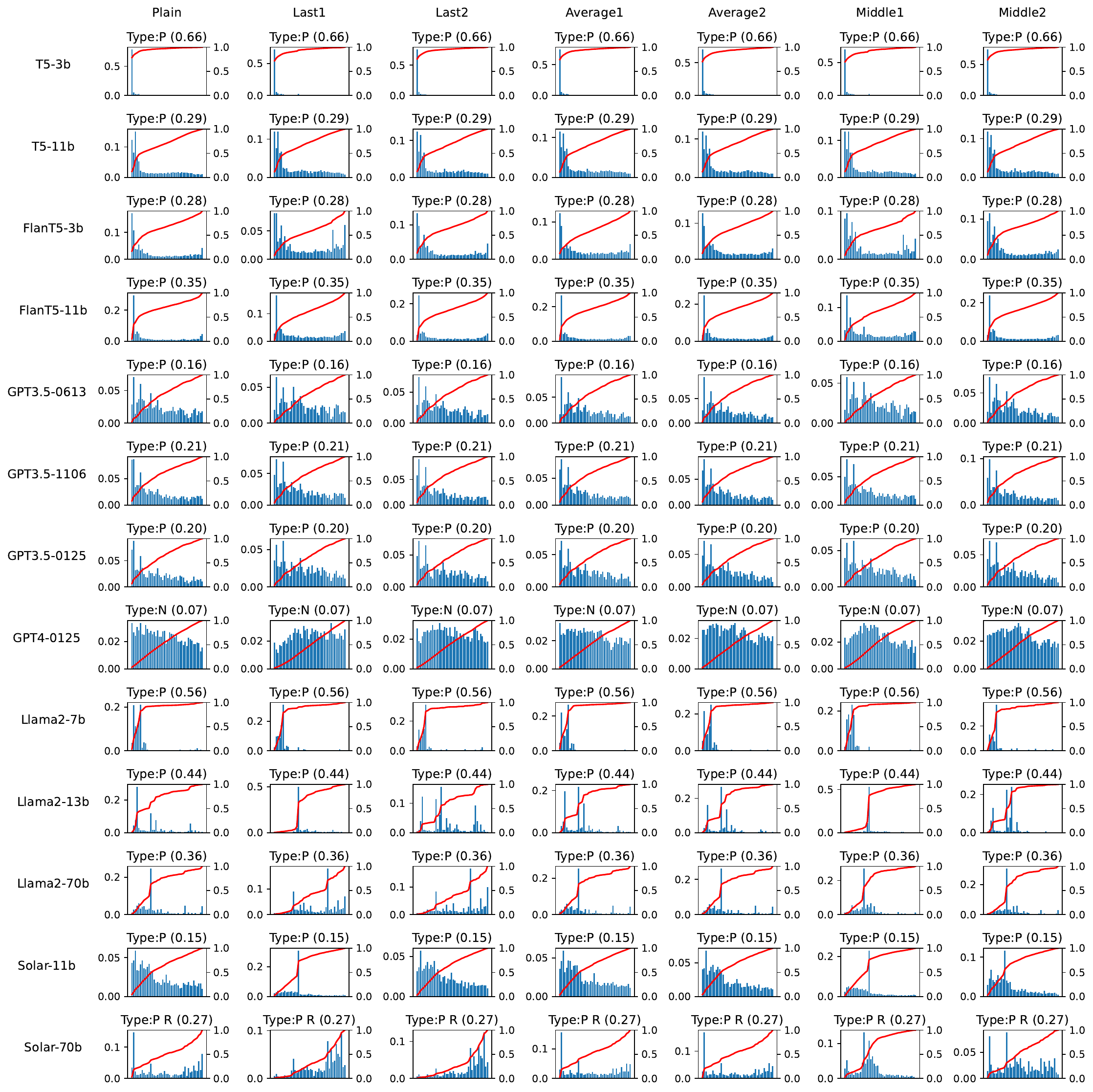}
        \caption{The label distribution of all models and prompts on the task of TACRED. The x-axis is the position of the labels. The y-axes are the probability of the label being chosen. The red line is the cumulative probability distribution. The \SPE{} type in labeled on the top of each distribution with the SPEM in the bracket.}
        \label{fig: tacred-distribution}
    \end{figure*}

\begin{figure*}[!hbt]
        \centering
        \includegraphics[width=1.9\columnwidth]{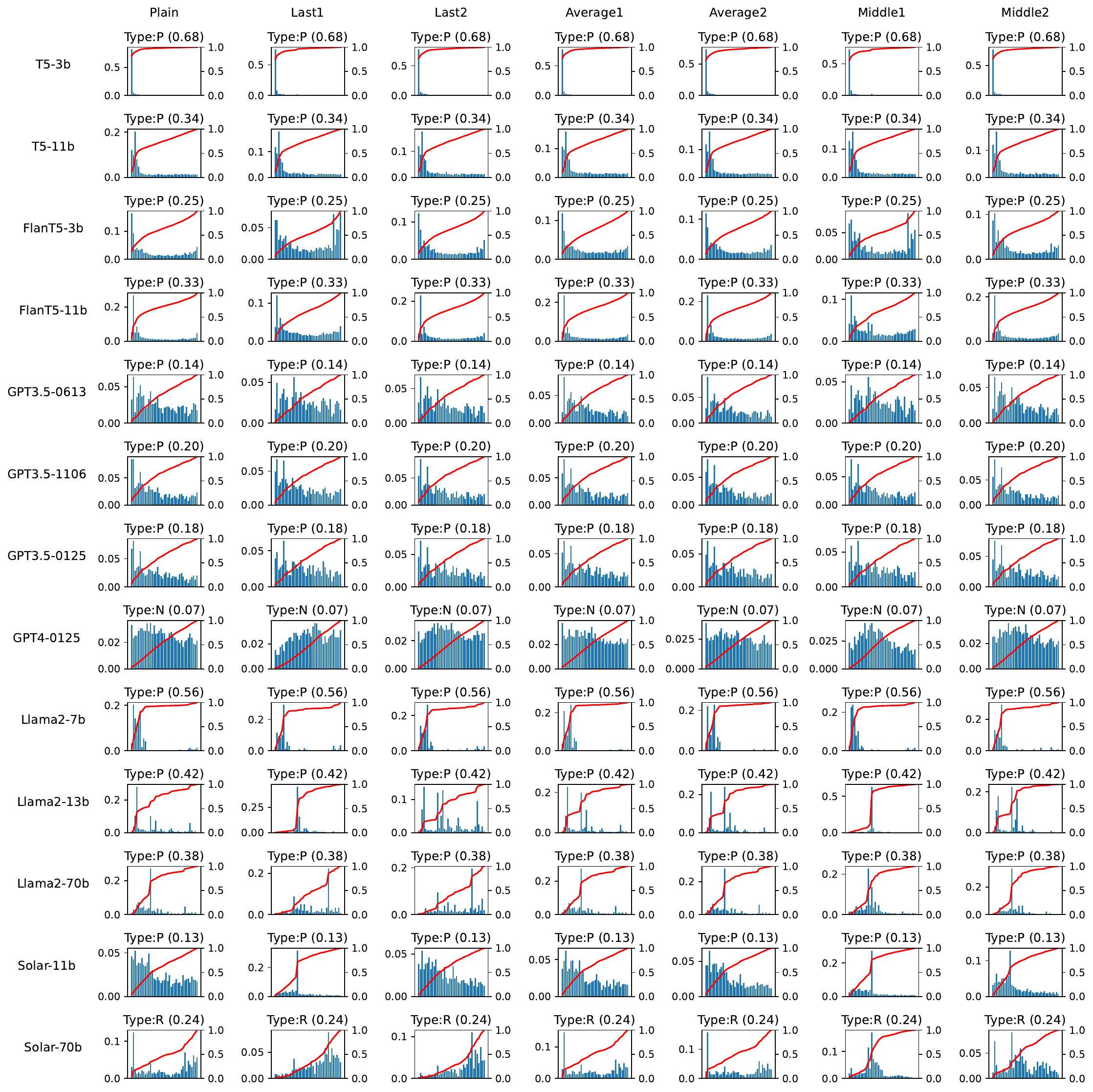}
        \caption{The label distribution of all models and prompts on the task of RETACRED. The x-axis is the position of the labels. The y-axes are the probability of the label being chosen. The red line is the cumulative probability distribution. The \SPE{} type in labeled on the top of each distribution with the SPEM in the bracket.}
        \label{fig: retacred-distribution}
    \end{figure*}

\begin{figure*}[!hbt]
        \centering
        \includegraphics[width=1.9\columnwidth]{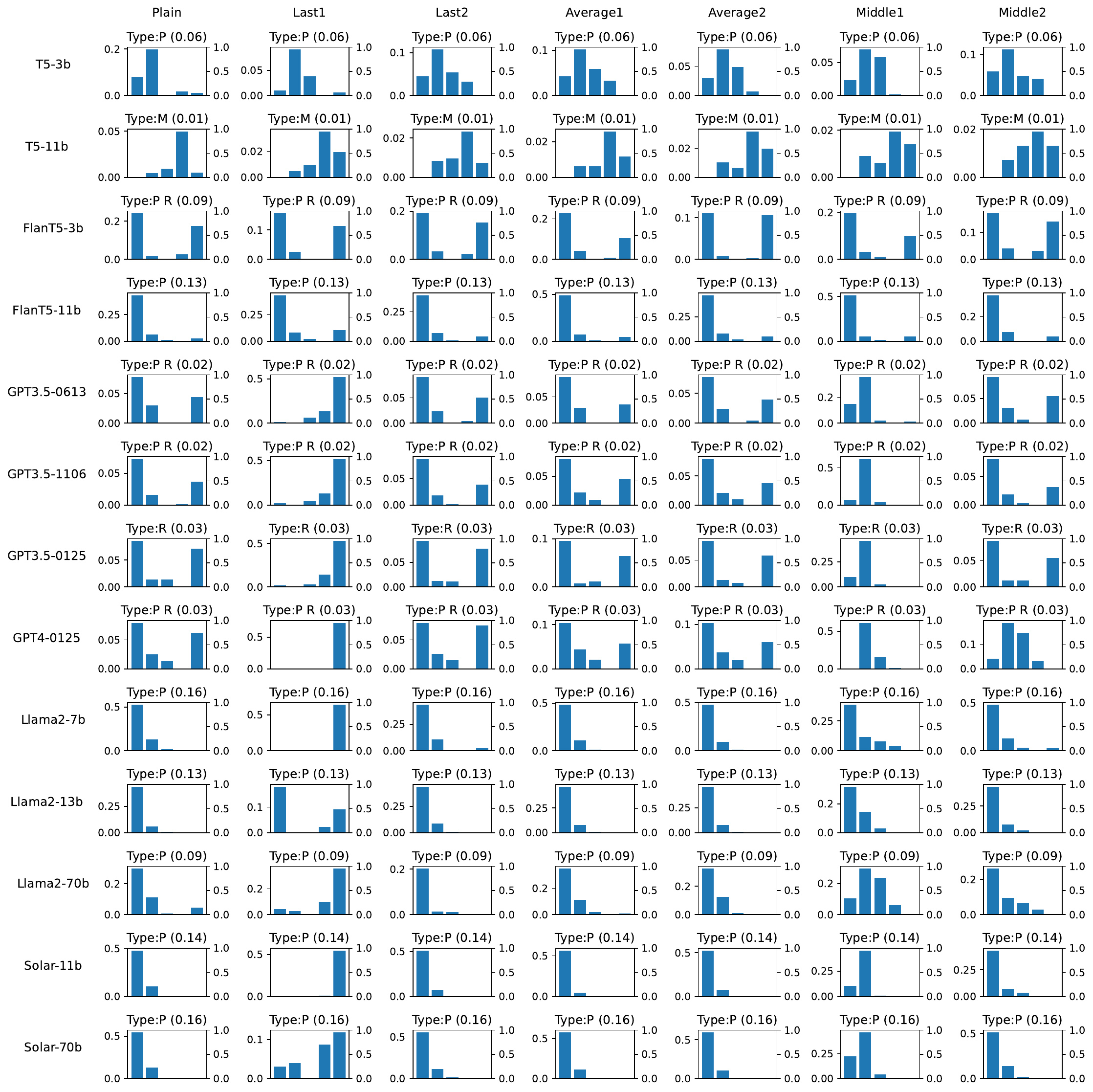}
        \caption{The BERTScore difference of all models and prompts on the task of Summ5. The x-axis is the position of the articles, the y-axes are the difference of BERTScores. The y-axes are the probability of the label being chosen. The \SPE{} type in labeled on the top of each distribution with the SPEM in the bracket.}
        \label{fig: summ5-distribution}
    \end{figure*}

\begin{figure*}[!hbt]
        \centering
        \includegraphics[width=1.9\columnwidth]{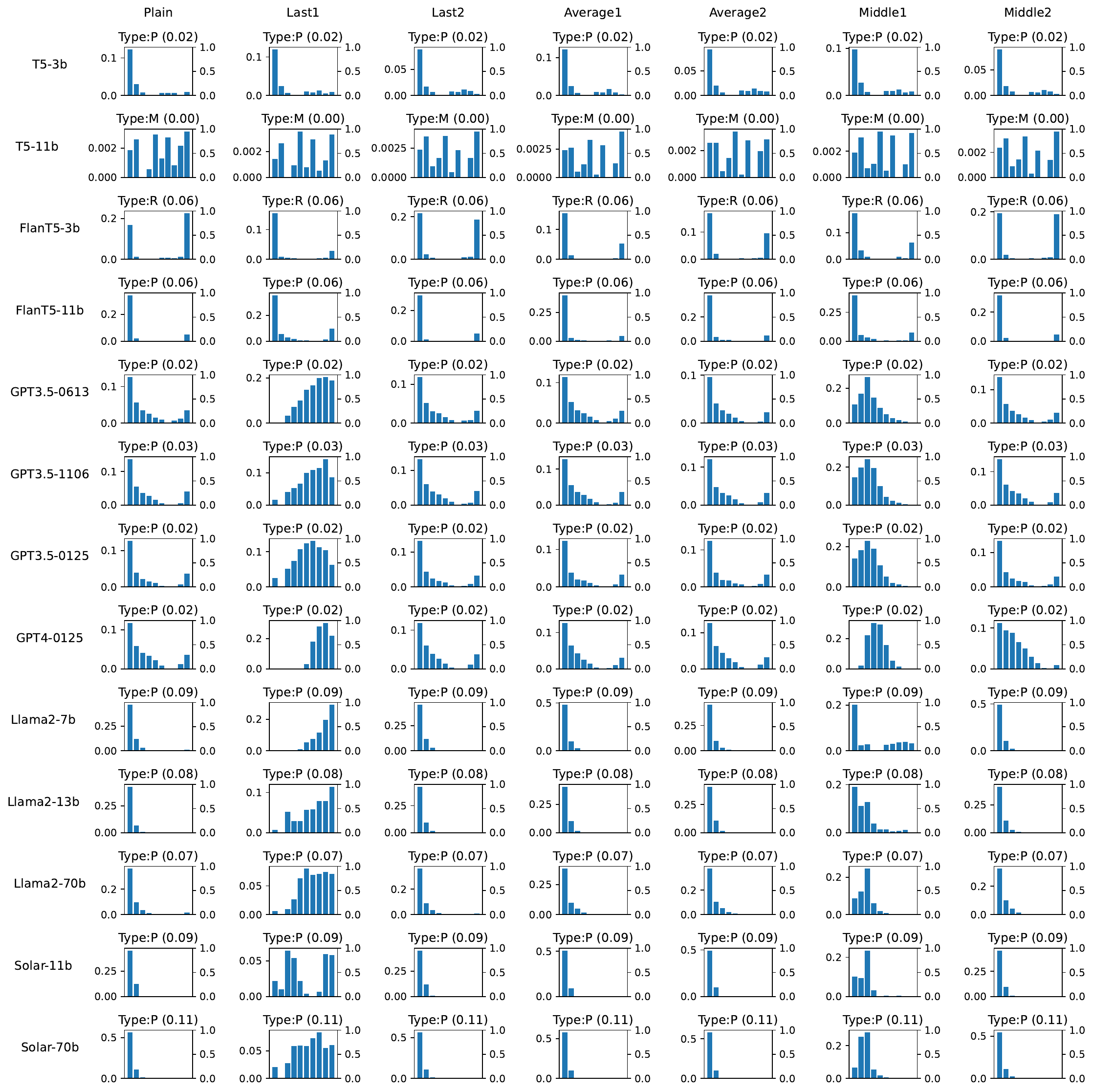}
        \caption{The BERTScore difference of all models and prompts on the task of Summ10. The x-axis is the position of the articles, the y-axes are the difference of BERTScores. The \SPE{} type in labeled on the top of each distribution with the SPEM in the bracket.}
        \label{fig: summ10-distribution}
    \end{figure*}

\begin{figure*}[!hbt]
        \centering
        \includegraphics[width=1.9\columnwidth]{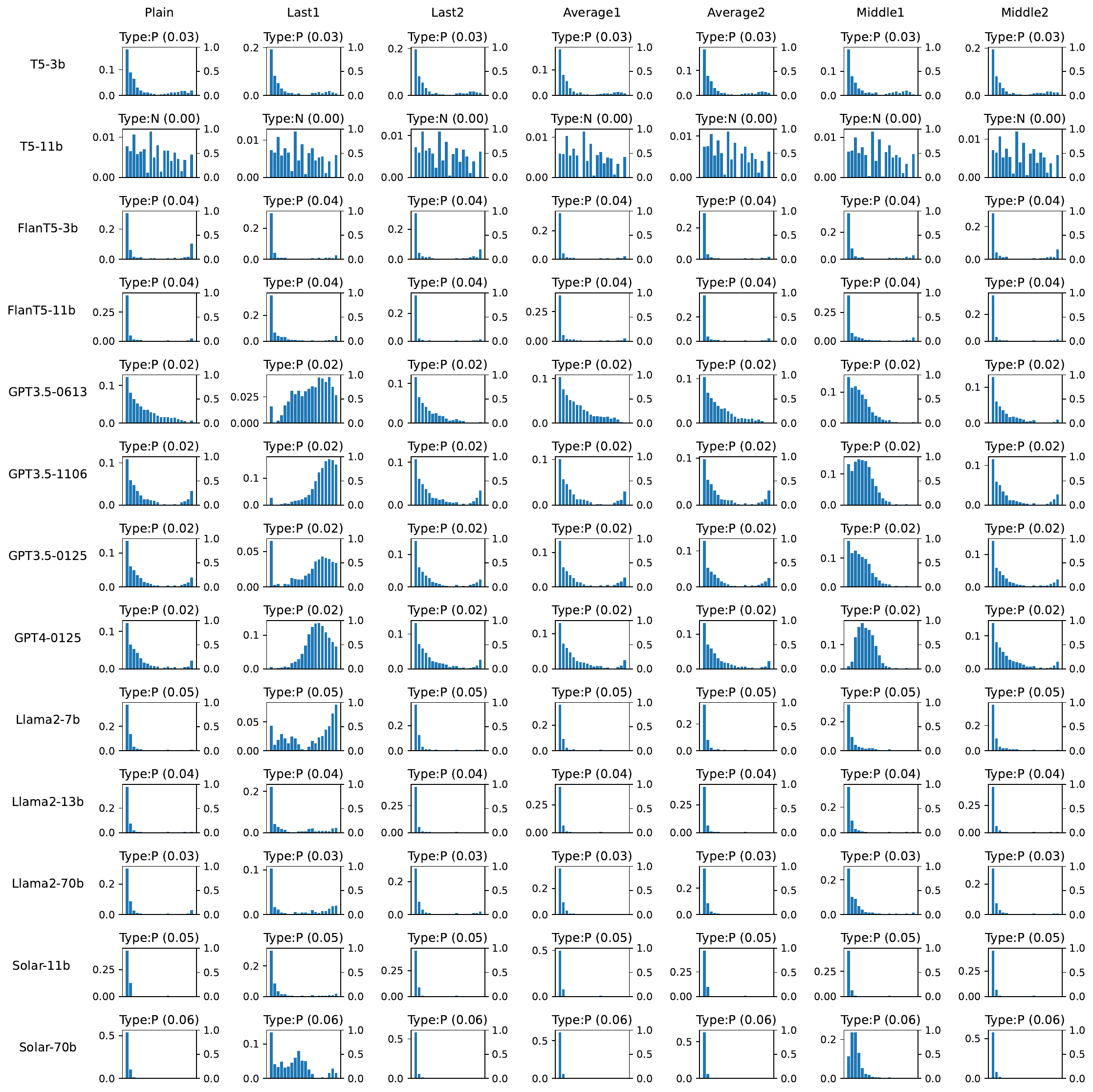}
        \caption{The BERTScore difference of all models and prompts on the task of Summ20. The x-axis is the position of the articles, the y-axes are the difference of BERTScores. The \SPE{} type in labeled on the top of each distribution with the SPEM in the bracket.}
        \label{fig: summ20-distribution}
    \end{figure*}

    In Figure~\ref{fig: banking77-distribution}, Figure~\ref{fig: goemotions-distribution}, Figure~\ref{fig: massive-distribution}, Figure~\ref{fig: tacred-distribution}, Figure~\ref{fig: retacred-distribution}, Figure~\ref{fig: summ5-distribution}, Figure~\ref{fig: summ10-distribution}, and Figure~\ref{fig: summ20-distribution}, we show the label distribution (for the label shuffling experiments) and BERTScore difference (for the summarization experiments) for all models with 7 different prompts on all tasks. These figures provide us an intuitive understand how the \SPE{} varies through tasks, models, and prompts.

\begin{figure}[!hbt]
        \centering
        \includegraphics[width=1.0\columnwidth]{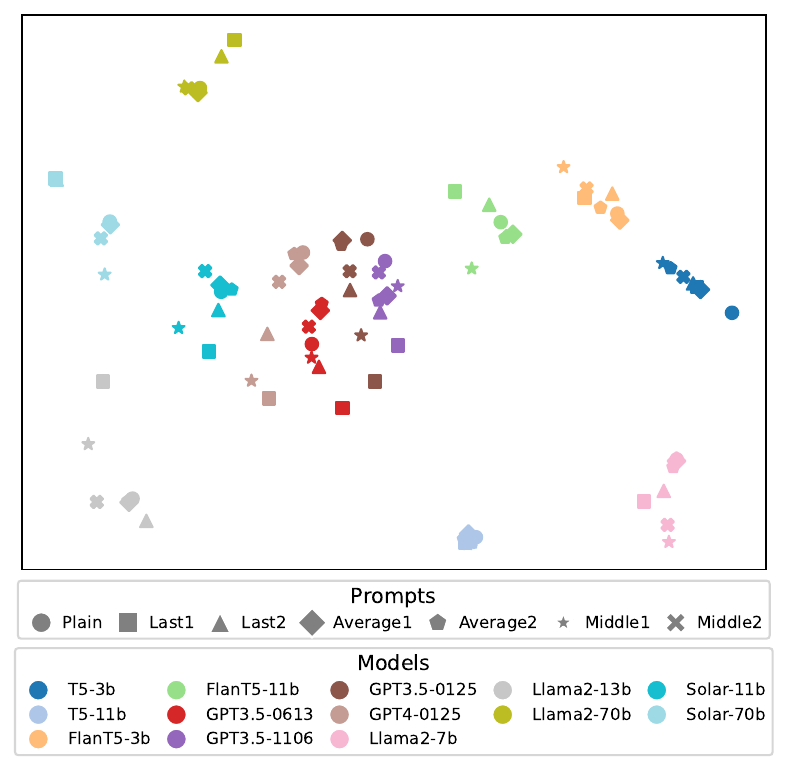}
        \caption{The t-SNE results of all models and prompts on the task of Banking77. The \SPE{} type in labeled on the top of each distribution with the SPEM in the bracket.}
        \label{fig: banking77-distribution-tsne}
    \end{figure}

\begin{figure}[!hbt]
        \centering
        \includegraphics[width=1.0\columnwidth]{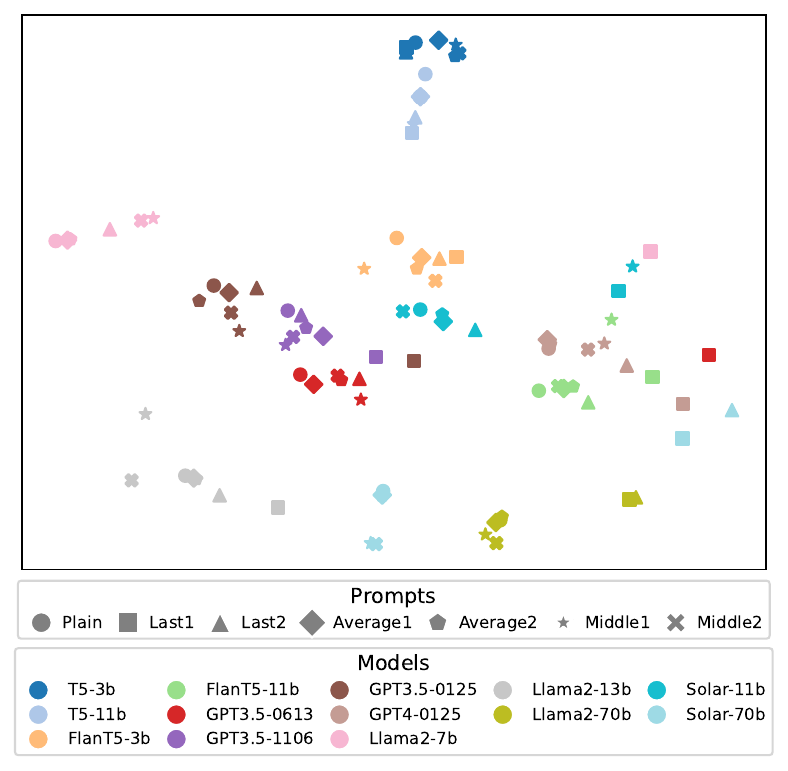}
        \caption{The t-SNE results of all models and prompts on the task of GoEmotions. The \SPE{} type in labeled on the top of each distribution with the SPEM in the bracket.}
        \label{fig: goemotions-distribution-tsne}
    \end{figure}

\begin{figure}[!hbt]
        \centering
        \includegraphics[width=1.0\columnwidth]{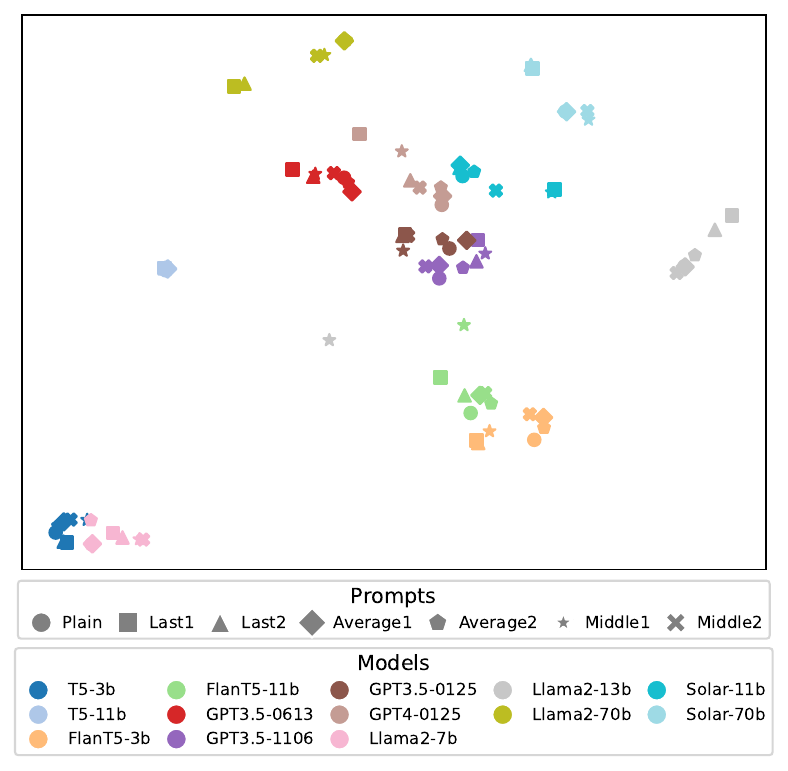}
        \caption{The t-SNE results of all models and prompts on the task of MASSIVE. The \SPE{} type in labeled on the top of each distribution with the SPEM in the bracket.}
        \label{fig: massive-distribution-tsne}
    \end{figure}

\begin{figure}[!hbt]
        \centering
        \includegraphics[width=1.0\columnwidth]{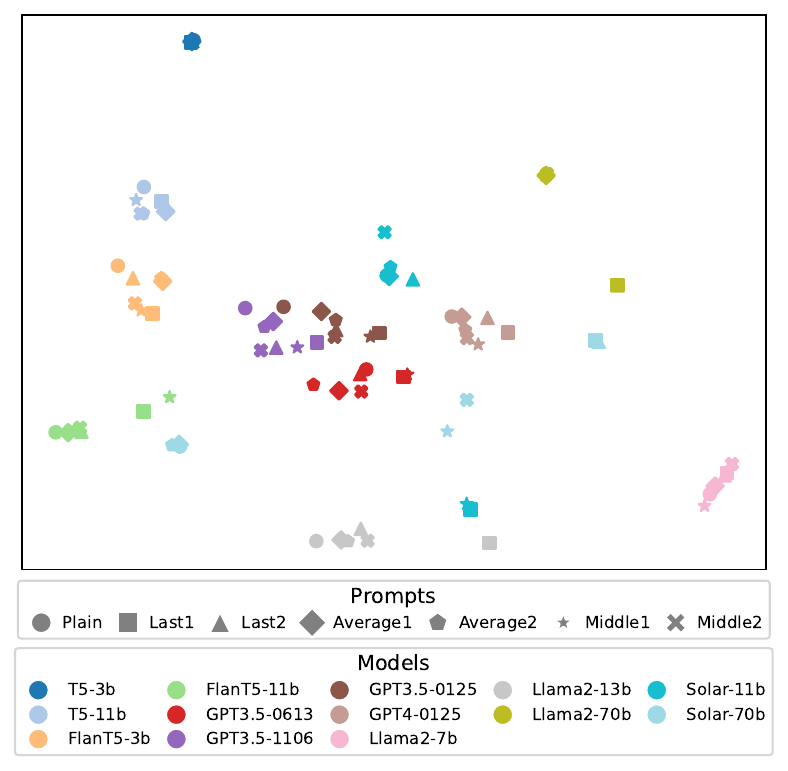}
        \caption{The t-SNE results of all models and prompts on the task of TACRED. The \SPE{} type in labeled on the top of each distribution with the SPEM in the bracket.}
        \label{fig: tacred-distribution-tsne}
    \end{figure}

\begin{figure}[!hbt]
        \centering
        \includegraphics[width=1.0\columnwidth]{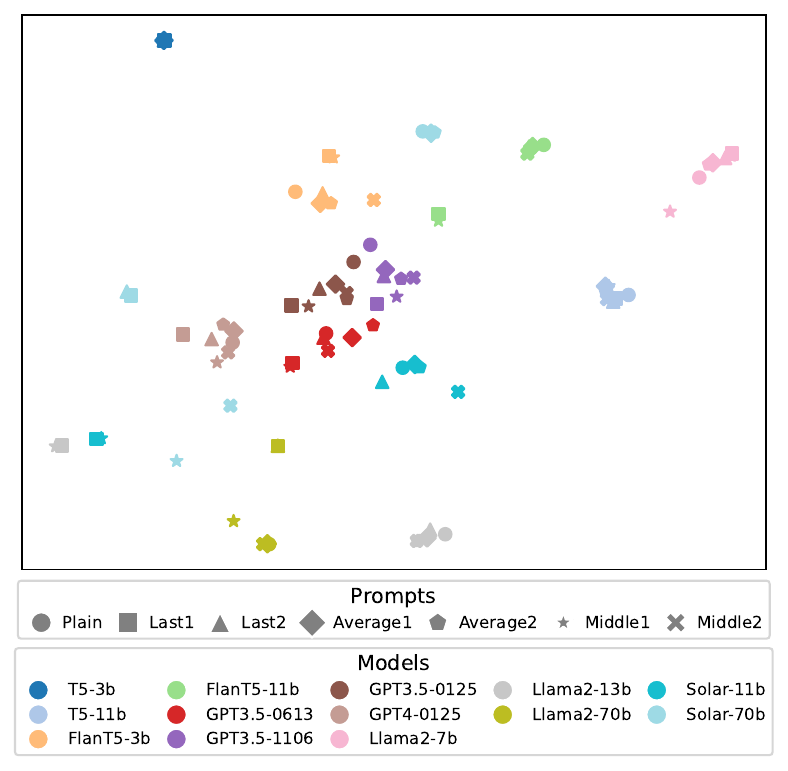}
        \caption{The t-SNE results of all models and prompts on the task of RETACRED. The \SPE{} type in labeled on the top of each distribution with the SPEM in the bracket.}
        \label{fig: retacred-distribution-tsne}
    \end{figure}

\begin{figure}[!hbt]
        \centering
        \includegraphics[width=1.0\columnwidth]{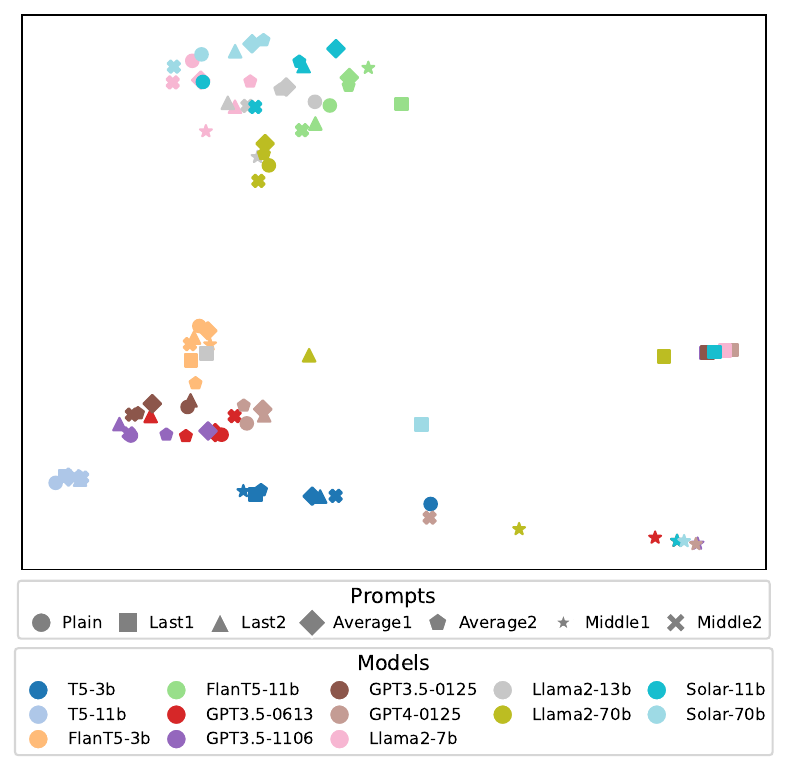}
        \caption{The t-SNE results of all models and prompts on the task of Summ5. The \SPE{} type in labeled on the top of each distribution with the SPEM in the bracket.}
        \label{fig: summ5-distribution-tsne}
    \end{figure}

\begin{figure}[!hbt]
        \centering
        \includegraphics[width=1.0\columnwidth]{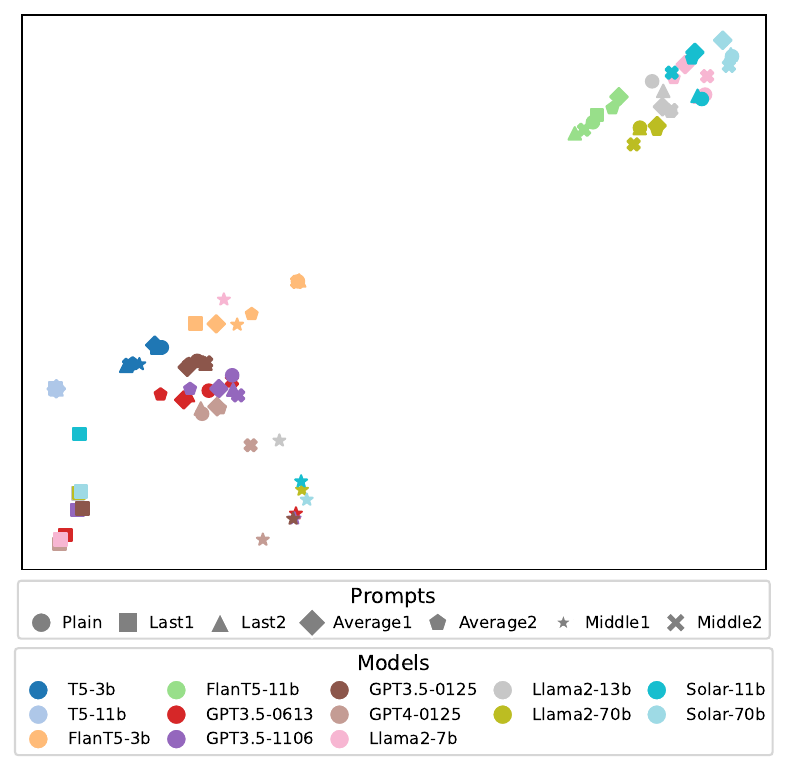}
        \caption{The t-SNE results of all models and prompts on the task of Summ10. The \SPE{} type in labeled on the top of each distribution with the SPEM in the bracket.}
        \label{fig: summ10-distribution-tsne}
    \end{figure}

\begin{figure}[!hbt]
        \centering
        \includegraphics[width=1.0\columnwidth]{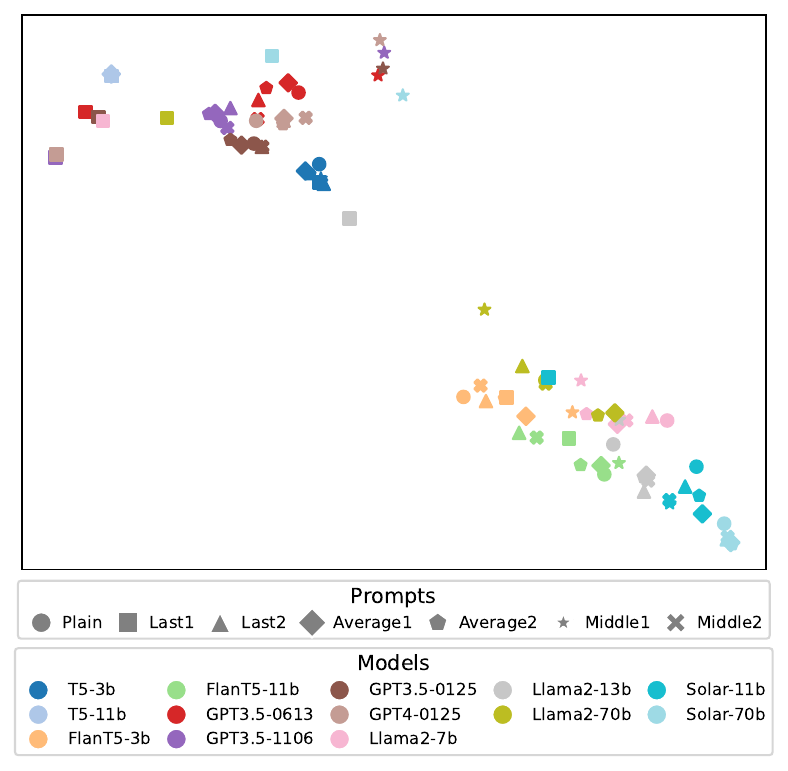}
        \caption{The t-SNE results of all models and prompts on the task of Summ20. The \SPE{} type in labeled on the top of each distribution with the SPEM in the bracket.}
        \label{fig: summ20-distribution-tsne}
    \end{figure}

    The t-SNE results for all datasets are shown in Figure Figure~\ref{fig: banking77-distribution-tsne}, Figure~\ref{fig: goemotions-distribution-tsne}, Figure~\ref{fig: massive-distribution-tsne}, Figure~\ref{fig: tacred-distribution-tsne}, Figure~\ref{fig: retacred-distribution-tsne}, Figure~\ref{fig: summ5-distribution-tsne}, Figure~\ref{fig: summ10-distribution-tsne}, and Figure~\ref{fig: summ20-distribution-tsne}.  Although the t-SNE results are different, we can still get the conclusion that the model architecture is main reason for the clustering. We also notice that for the summarization task, the Prompt might be of more evident influence to the \SPE{} such as the ``Middle1'' of Summ5 in Figure~\ref{fig: summ10-distribution-tsne} and ``Last1'' of Summ20 in Figure~\ref{fig: summ20-distribution-tsne}.


\end{document}